\pdfoutput=1

\documentclass[11pt]{article}

\usepackage{acl}

\usepackage{times}
\usepackage{latexsym}
\usepackage{multirow}
\usepackage{colortbl}
\usepackage{makecell}

\usepackage[T1]{fontenc}

\usepackage[utf8]{inputenc}

\usepackage{microtype}

\usepackage{inconsolata}

\usepackage{amsfonts}
\usepackage{graphicx}
\usepackage{color}
\usepackage{multicol}
\usepackage{multirow}
\usepackage{amsmath} 

\title{Advancing Parameter Efficiency in Fine-tuning via Representation Editing}

\author{Muling Wu, Wenhao Liu, Xiaohua Wang, Tianlong Li, Changze Lv \\
{\bf Zixuan Ling, Jianhao Zhu, Cenyuan Zhang,  Xiaoqing Zheng\thanks{Corresponding author.}, Xuanjing Huang} \\
  School of Computer Science, Fudan University, Shanghai, China \\
  \texttt{\{mlwu22,whliu22,xiaohuawang22,tlli22,czlv22\}@m.fudan.edu.cn} \\
 \texttt{\{zhengxq,xjhuang\}@fudan.edu.cn} \\}

\begin{document}
\maketitle
\begin{abstract}

Parameter Efficient Fine-Tuning (PEFT) techniques have drawn significant attention due to their ability to yield competitive results while updating only a small portion of the adjustable parameters. 
However, existing PEFT methods pose challenges in hyperparameter selection, such as choosing the rank for LoRA or Adapter, or specifying the length of soft prompts. 
To address these challenges, we propose a novel fine-tuning approach for neural models, named Representation EDiting (RED), which modifies the representations generated at some layers through the application of scaling and biasing operations. 
While existing PEFT methods still demonstrate over-parameterization that could potentially undermine the generalization ability acquired from pre-training, RED can substantially reduce the number of trainable parameters by a factor of $25,700$ compared to full parameter fine-tuning and by a factor of $32$ relative to LoRA. Remarkably, RED achieves results comparable or superior to both full parameter fine-tuning and other PEFT methods. Extensive experiments across various model architectures and scales, including RoBERTa, GPT-2, T5, and LLaMA-2, have demonstrated the effectiveness and efficiency of RED\footnote{The code is accessible at \url{https://github.com/mlwu22/RED}.}, thereby positioning it as a promising PEFT strategy for large-scale neural models.

\end{abstract}
\section{Introduction}


Pre-training on large-scale unlabeled datasets followed by fine-tuning on task-specific dataset has demonstrated remarkable efficacy across various natural language processing (NLP) tasks, establishing itself as the prevailing training paradigm \cite{devlin2018bert,raffel2020exploring,radford2018improving}. However, conducting full parameter fine-tuning for each task can be exceedingly costly and increasingly daunting as model scales continue to grow \cite{brown2020language}. For instance, BERT comprises up to $220$ million parameters, T5 scales up to $11$ billion parameters, and GPT-3 boasts an astounding $175$ billion parameters. Consequently, the efficient and effective adaptation of large models to specific downstream tasks presents an intriguing research challenge \cite{He2021TowardsAU}.


In response to this challenge, researchers have put forward three main lines of Parameter Efficient Fine-Tuning (PEFT) techniques \cite{ding2022delta}.
Firstly, addition-based methods involve the introduction of additional trainable neural modules or parameters that were not present in the original model \cite{houlsby2019parameter,karimi2021compacter,li2021prefix,lester2021power}.
Specification-based methods, on the other hand, identify certain parameters in the original model to be trainable, while the rest are kept frozen \cite{zaken2021bitfit,guo2020parameter}.
Lastly, reparameterization-based methods reconfigure trainable parameters into a more parameter-efficient form through certain transformations \cite{hu2021lora,zhang2023adaptive,ding2023sparse}.


Among these PEFT methods, Low-Rank Adaptation (LoRA) stands out as one of the most efficient techniques with its effectiveness empirically validated across various models of diverse scales. 
Despite its impressive performance, LoRA still demands a significant number of trainable parameters. 
Recent studies by \citet{aghajanyan2020intrinsic} and \citet{Kopiczko2023VeRAVR} indicate that the upper bound for intrinsic dimensions is substantially smaller than what is typically used in such methods. For example, the $d_{90}$ value (the minimum number of trainable parameters required to reach $90\%$ of the performance of the fully-parameter fine-tuned model) for the RoBERTa base is reported to be $896$. However, when LoRA is applied to fine-tune this model, the number of trainable parameters escalates to $0.3$ million, suggesting the potential for further reduction in parameter count.


In addition to the issue of requiring too many adjustable parameters, existing PEFT methods \cite{Mao2021UniPELTAU,He2021TowardsAU,ding2022delta} primarily focus on the design of lightweight modules and their integration (or placement) within base models. 
Nonetheless, the implementation of these PEFT techniques introduces additional complexities in hyperparameter selection, such as choosing the rank of LoRA and Adapter, or deciding on the length of Soft Prompt and Prefix.

\begin{figure*}
    \centering
\noindent\includegraphics[width=0.82\textwidth]{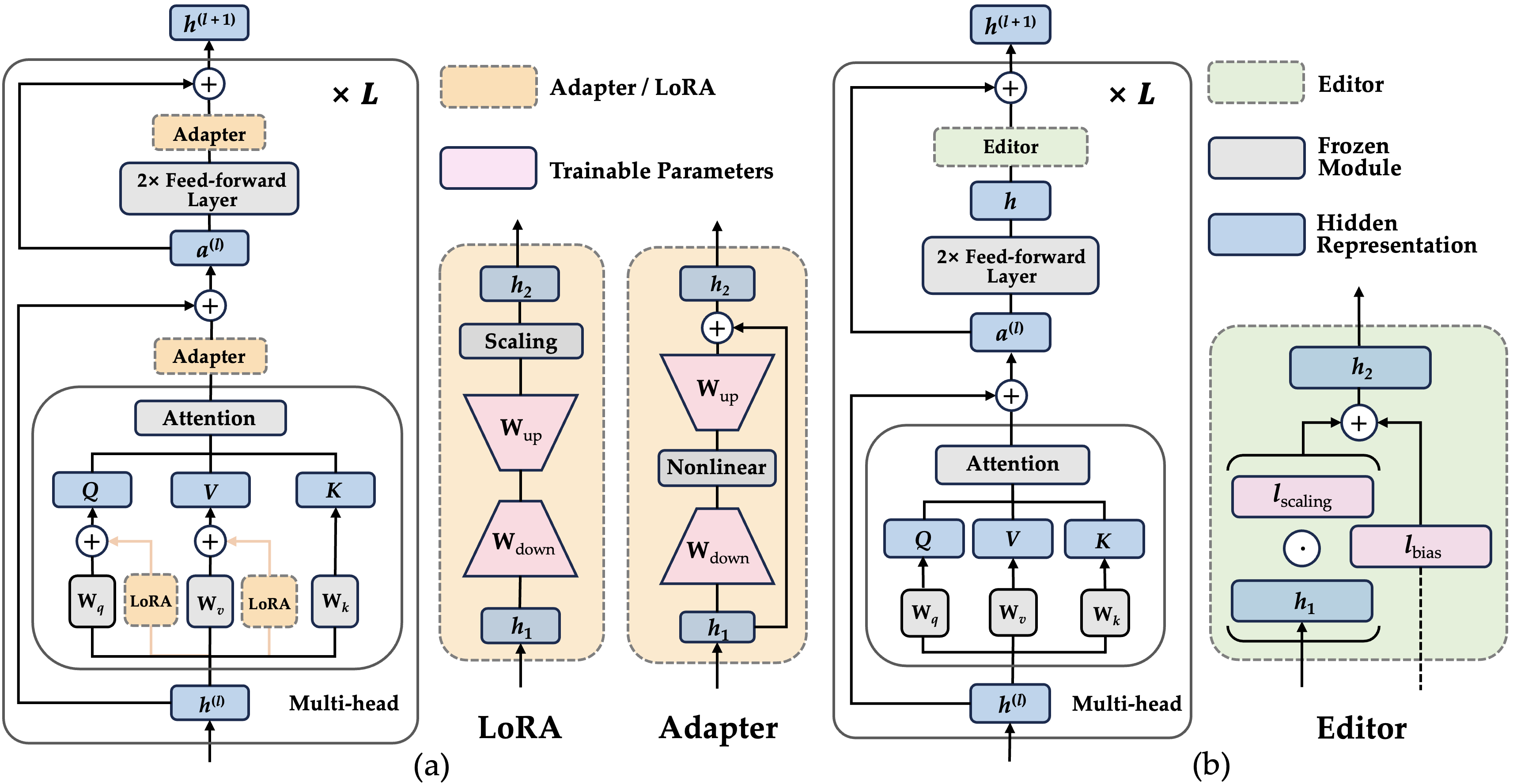}
\vspace{-3mm}
    \caption{\label{fig:figure_01}
Comparison of previous representative PEFT methods with the proposed RED.
(a) LoRA incorporates learnable bottleneck-shaped modules (highlighted in \textcolor{orange}{orange}) by integrating additional connections parallel to the $\mathbf{W}_q$ and $\mathbf{W}_v$ matrices of attention blocks, along with modifying the weights of these matrices in a low-rank fashion. 
Adapter, on the other hand, introduces learnable modules within similar structures (also highlighted in \textcolor{orange}{orange}) by incorporating additional connections following both the attention and feed-forward sub-layers.
(b) RED introduces two learnable vectors, $l_\text{scaling}$ and $l_\text{bias}$, to directly edit the representations (marked in \textcolor{teal}{green}) generated by feed-forward sub-layers, which significantly reduces the number of parameters required for fine-tuning.
\vspace{-5mm}
}
    
\end{figure*}

Inspired by the concept of representation engineering \cite{Zou2023RepresentationEA}, we shift our focus away from the weights of models and turn our attention to their representations.
In the neural architecture, network weights govern neural activities (or representations), which in turn determine the networks' output, and the networks' output ultimately shapes the networks' behavior. Rather than concentrating on neurons and their interconnections (or weights), we explored to achieve control over network behavior by manipulating its internal representations. Specifically, we fine-tune neural network models by directly editing the representation generated at each layer while maintaining the model parameters frozen, as illustrated in Figure \ref{fig:figure_01} (b).
It is worth noting that the number of parameters required to edit representations is substantially fewer than that of weights within neural networks.
Taking LLaMA-2 ($7$B) as an example, the proposed representation editing (RED) method achieves competitive performance by adjusting only $0.26$M parameters. 
This is approximately $25,700$ times less than what is required for full-parameter fine-tuning, rendering the method both storage and computation efficient.



The contributions of this study are summarized as follows:
\begin{itemize}
\setlength{\itemsep}{0pt}
\setlength{\parsep}{0pt}
\setlength{\parskip}{0pt}
\item We propose a novel perspective on fine-tuning by directly editing model representations, diverging from exiting PEFT methods that focused on adjusting the model's weights. Our proposed PEFT technique, termed RED, embodies this new perspective.
\item Extensive experiments are conducted across models of varying structures and scales, including RoBERTa, GPT-2, T5, and LLaMA-2. 
The effectiveness of RED is validated across a range of natural language understanding and generation tasks.
Notably, RED demonstrates both efficacy and efficiency while requiring only a minimal number of trainable parameters and maintaining ease of implementation.
\item A comprehensive ablation study is conducted to dissect the individual components of RED and understand their impacts on performance.
\end{itemize}


\section{Related Work}

Existing PEFT methods can be broadly classified into three categories \cite{ding2022delta}, each characterized by the particular parts of parameters that they tune efficiently: addition-based, specification-based, and reparameterization-based methods.


Addition-based methods perform fine-tuning by integrating additional lightweight learnable components into foundational models. More specifically, \citet{houlsby2019parameter,stickland2019bert,karimi2021compacter} and \citet{Rckl2020AdapterDropOT} proposed the integration of learnable bottleneck neural modules into the transformer layers. \citet{brown2020language} and \citet{Shin2020ElicitingKF} found that the models' performance could be enhanced by appending discrete tokens (or prompts) at the beginning of the input texts without modifying the models' parameters. However, the manual creation of such prompts demands substantial effort, and optimization in a discrete space is comparatively more challenging. Consequently, later studies substituted these discrete tokens with continuous vectors, also referred to as soft prompts, which can be optimized using the gradient descent algorithm \cite{Lester2021ThePO, Li2021PrefixTuningOC, Wu2023ParameterEM, Wang2023MultitaskPT}.

Specification-based methods achieve parameter-efficient fine-tuning by designating a subset of parameters for modification while leaving the remaining parameters untouched. Within this category, \citet{Lee2019WhatWE} suggested a method to exclusively update the parameters in certain top layers of BERT and RoBERTa.
BitFit \cite{BenZaken2021BitFitSP}, on the other hand, fine-tunes a model by only optimizing the bias terms used in the model. Contrary to these methods that pre-determine which parameters should be altered, \citet{guo2020parameter} and \citet{Zhao2020MaskingAA} implemented a learnable masking strategy to automatically choose the parameters for tuning.

Reparameterization-based methods aim to optimize some parameters within a model, typically in their low-dimensional subspace.
\citet{hu2021lora} introduces the use of low-rank matrices, termed LoRA, to approximate the weight modifications during the fine-tuning process.
QLoRA \cite{Dettmers2023QLoRAEF} combines low-rank adaptation with model quantization to further reduce storage usage during the fine-tuning phase.
AdaLoRA \cite{Zhang2023AdaptiveBA} proposes to use SVD decomposition to approximate the changes in weights, which allocates more trainable parameters to more important weight matrices, leading to a better performance.



\citet{Zou2023RepresentationEA} proposed an approach to analyzing and manipulating the behavior of neural networks through representation engineering, thereby shifting the focus from neurons and their circuits to representations and the transformations among them.
\citet{Liu2023AligningLL} extended this work to align large language models with human preferences, achieving results comparable or superior to RLHF \cite{ouyang2022training}, but at a lower computational cost.
\citet{Subramani2022ExtractingLS} investigated the extraction of ``steer vectors'' from the hidden layers and performed unsupervised text style transfer by altering the hidden representations using these vectors.
To control the style and sentiment expressed in a model's outputs, \citet{Turner2023ActivationAS} integrated a similar ``steer vector'' into the representation of each hidden layer at the inference time.
Inspired by the recent advance in representation engineering \citet{Zou2023RepresentationEA}, we suggested a novel perspective for fine-tuning models by directly editing their representations. This new perspective was embodied in the RED training method, which demonstrated both effectiveness and efficiency while requiring only a minimal number of trainable parameters.


This study is closely related to two recent works: IA3 \cite{Liu2022FewShotPF} and SSF \cite{Lian2022ScalingS}. 
To fulfill the in-context learning capability with lower computational costs, \citet{Liu2022FewShotPF} proposed to modify the key and value vectors of the multi-head attentions and those generated by feed-forward networks (FFNs) by scaling operations. 
In the domain of computer vision, \citet{Lian2022ScalingS} presented a similar method, called SSF, which was motivated by batch and layer normalization techniques.
To ensure features fall in a discriminative space for better classification, SSF also necessitates the adjustment of feature vectors across nearly all transformer layers, including multi-head attentions, FFNs, and normalization layers. 
As a result, the size of parameters adjusted in their method approximates those manipulated in other PEFT techniques, such as Adapter and VPT \cite{jia2022visual}. 

In contrast, our RED method requires only the modification of activation patterns generated by FFN sub-layers, constituting a significantly smaller portion of the entire neural network's parameters for model fine-tuning. 
While SSF achieves a reduction in the number of fine-tuning parameters by a factor of $300$ compared to original models, RED accomplishes a reduction of approximately $25,700$ times. Moreover, SSF's evaluation was restricted to a limited range of simple image classification tasks using relatively small models (under $100$M parameters), whereas RED has demonstrated its effectiveness across a variety of natural language understanding and generation tasks, with significantly larger models of up to $7$B parameters.

\begin{table*}[t] 
\small
\centering
\setlength{\tabcolsep}{5pt}
\begin{tabular}{l|r|cccccccc|c}
\Xhline{1pt}
\textbf{Method} & \textbf{\# Param} & \textbf{MNLI} & \textbf{SST-2} & \textbf{MRPC} & \textbf{CoLA} & \textbf{QNLI} & \textbf{QQP} & \textbf{RTE} & \textbf{STS-B} & \textbf{Avg.} \\
\Xhline{1pt}
\textbf{FT (base)} & $125\mathbf{M}$ & $87.3$ & $94.4$ & $87.9$ & $62.4$ & $92.5$ & $91.7$ & $78.3$ & $90.6$ & $85.6$ \\
\textbf{Adapter (base)} & $0.4\mathbf{M}$ & $87.0$ & $93.3$ & $88.4$ & $60.9$ & $92.5$ & $90.5$ & $76.5$ & $90.5$ & $85.0$ \\
\textbf{LoRA (base)} & $0.3\mathbf{M}$ & $86.6$ & $93.9$ & $88.7$ & $59.7$ & $92.6$ & $90.4$ & $75.3$ & $90.3$ & $84.7$ \\
\textbf{Adapter-FFN (base)} & $0.3\mathbf{M}$ & $87.1$ & $93.0$ & $88.8$ & $58.5$ & $92.0$ & $90.2$ & $77.7$ & $90.4$ & $84.7$ \\
\textbf{BitFit (base)} & $0.1\mathbf{M}$ & $84.7$ & $94.0$ & $88.1$ & $54.0$ & $91.0$ & $87.3$ & $69.8$ & $89.5$ & $82.3$ \\
\textbf{IA3 (base)} & $0.06\mathbf{M}$ & $85.4$ & $93.4$ & $86.4$ & $57.8$ & $91.1$ & $88.5$ & $73.5$ & $88.5$ & $83.1$ \\
\rowcolor{gray!30}
\textbf{RED (base)} & $0.02 \mathbf{M}$ & $83.9$ & $93.9$ & $89.2$ & $61.0$ & $90.7$ & $87.2$ & $78.0$ & $90.4$ & $84.3$ \\
\hline
\textbf{FT (large)} & $355\mathbf{M}$ & $88.8$ & $96.0$ & $91.7$ & $68.2$ & $93.8$ & $91.5$ & $85.8$ & $92.6$ & $88.5$ \\
\textbf{Adapter (large)} & $0.9\mathbf{M}$ & $90.1$ & $95.2$ & $90.5$ & $65.4$ & $94.6$ & $91.4$ & $85.3$ & $91.5$ & $88.0$ \\
\textbf{LoRA (large)} & $0.8\mathbf{M}$ & $90.2$ & $96.0$ & $89.8$ & $65.5$ & $94.7$ & $90.7$ & $86.3$ & $91.7$ & $88.1$ \\
\textbf{Adapter-FFN (large)} & $0.8\mathbf{M}$ & $90.3$ & $96.1$ & $90.5$ & $64.4$ & $94.3$ & $91.3$ & $84.8$ & $90.2$ & $87.7$ \\
\textbf{IA3 (large)} & $0.15\mathbf{M}$ & $90.1$ & $94.5$ & $87.1$ & $63.2$ & $93.9$ & $89.3$ & $85.3$ & $91.5$ & $86.9$ \\
\rowcolor{gray!30}
\textbf{RED (large)} & $0.05\mathbf{M}$ & $89.5$ & $96.0$ & $90.3$ & $68.1$ & $93.5$ & $88.8$ & $86.2$ & $91.3$ & $87.9$ \\
\Xhline{1pt}
\end{tabular}

\caption{Performance comparison of RoBERTa base and large models fine-tuned by RED and other PEFT baseline methods on the GLUE benchmark. 
The standard deviations of results across different methods are given in Table \ref{tab:standaerd_appendix_roberta}.}
\label{tab:roberta}
\vspace{-5mm}
\end{table*}

\section{Method}

We in this section give a concise overview of previous PEFT methods and introduce a new, parameter-efficient fine-tuning method known as Representation Editing (RED) by highlighting its distinctive features compared to existing PEFT methods.
RED facilitates the adaptation of pre-trained neural network models to downstream tasks by directly editing the model's representations.

\subsection{Recap of PEFT Methods}

The predominant large language models (LLMs) have been constructed based on the transformer architecture proposed in a seminal work on attention mechanisms \cite{Vaswani2017AttentionIA}.
This architecture is characterized by a series of layered blocks, each incorporating two fundamental sub-layers: multi-head self-attention and fully connected feed-forward networks.
Many existing PEFT methods usually achieve fine-tuning by adjusting the parameters of these two sub-layers in some parameter-efficient manner.
Figure \ref{fig:figure_01} (a) illustrates two exemplar PEFT methods: Adapter and LoRA. 
With the exception of a few additional parameters that require tuning, the parameters of the pre-trained LLM remain fixed.

Specifically, LoRA \cite{hu2021lora} incorporates learnable bottleneck-shaped modules through the connections parallel to the $\mathbf{W}_{q}$ and $\mathbf{W}_v$ matrices of attention blocks, as illustrated in Figure \ref{fig:figure_01} (a), and models the weight changes of these two matrices in a low-rank manner. 
Given a pre-trained weight matrix $\mathbf{W} \in \mathbb{R}^{d \times k}$, LoRA represents its changes, denoted as $\Delta \mathbf{W}$, through two low-rank decomposition matrices: $\Delta \mathbf{W} = \alpha \cdot \mathbf{W}_\text{down} \mathbf{W}_\text{up}$, where $\mathbf{W}_\text{down} \in \mathbb{R}^{d \times r}$ and $\mathbf{W}_\text{up} \in \mathbb{R}^{r \times k}$ (the value of $r$ is typically much smaller than both $d$ and $k$). 
The coefficient $\alpha$ is a hyperparameter that determines the significance of $\Delta \mathbf{W}$ relative to the original $\mathbf{W}$.
Given an input $x$, LoRA produces the result of the forward pass as follows:
\begin{equation} \label{eq:lora}
{h} = {x}\mathbf{W}+ \alpha \cdot {x} \mathbf{W}_\text{down}\mathbf{W}_\text{up}
\end{equation}

In the initial version of the Adapter, \citet{houlsby2019parameter} suggested the insertion of trainable adapter modules between two sub-layers within the transformer architecture.
These adapter modules are structured to include a down-projection matrix, denoted as $\mathbf{W}_\text{down} \in \mathbb{R}^{d \times r}$, which transforms a hidden representation ${h}_{1} \in \mathbb{R}^{d}$ to a lower-dimensional space with a pre-specified dimensionality $r$.
Subsequently, this dimensionally reduced vector undergoes a nonlinear activation function $f (\cdot)$ and an additional up-projection matrix $\mathbf{W}_\text{up} \in \mathbb{R}^{r \times d}$ to revert it to its original dimension $d$. 
The adapter module also incorporates a residual connection. 
The resultant output ${h}_{2} \in \mathbb{R}^{d}$ generated by this module can be formally represented as follows:
\begin{equation} \label{eq:adapter}
{h}_{2}={h}_{1}+ f({h}_{1}\mathbf{W}_\text{down})\mathbf{W}_\text{up}
\end{equation}

Expanding on this research, \citet{Pfeiffer2020AdapterFusionNT} proposed a more efficient variant of the Adapter that is only applied following the FFN sub-layer.

\begin{table*}[t] 
\small
\centering
\setlength{\tabcolsep}{8pt}
\begin{tabular}{l|r|ccccc}
\Xhline{1pt}
\textbf{Method} & \textbf{\# Param} & \textbf{BLEU} & \textbf{NIST} & \textbf{MET} & \textbf{ROUGE-L} & \textbf{CIDEr} \\
\Xhline{1pt}
\textbf{FT (medium)} & $355\mathbf{M}$ & $65.95$ & $8.52$ & $45.95$ & $69.13$ & $2.35$  \\
\textbf{$\mathbf{FT^{top2}}$ (medium)} & $25.2\mathbf{M}$ & $65.94$ & $8.53$ & $44.28$ & $68.83$ & $2.23$ \\
\hline 
\textbf{Adapter (medium)} & $0.9\mathbf{M}$ & $64.31$ & $8.29$ & $44.91$ & $67.72$ & $2.28$  \\
\textbf{LoRA (medium)} & $0.8\mathbf{M}$ & $67.43$ & $8.65$ & $46.01$ & $69.64$ & $2.42$  \\
\textbf{Adapter-FFN (medium)} & $0.8\mathbf{M}$ & $64.41$ & $8.30$ & $44.74$ & $67.53$ & $2.29$ \\
\textbf{Prefix Tuning (medium)} & $0.8\mathbf{M}$ & $63.92$ & $8.26$ & $41.81$ & $66.86$ & $2.03$  \\
\textbf{IA3 (medium)} & $0.17\mathbf{M}$ & $63.63$ & $7.99$ & $40.49$ & $66.36$ & $1.89$  \\
\rowcolor{gray!30}
\textbf{RED (medium)} & $0.05\mathbf{M}$ & $64.86$ & $8.36$ & $44.99$ & $67.62$ & $2.28$  \\
\hline
\textbf{FT (large)} & $774\mathbf{M}$ & $65.56$ & $8.50$ & $45.40$ & $68.38$ & $2.27$  \\
\textbf{Adapter (large)} & $1.8\mathbf{M}$ & $65.94$ & $8.46$ & $45.78$ & $68.65$ & $2.34$  \\
\textbf{LoRA (large)} & $1.5\mathbf{M}$ & $68.24$ & $8.76$ & $46.23$ & $69.92$ & $2.42$  \\
\textbf{Adapter-FFN (large)} & $1.5\mathbf{M}$ & $65.53$ & $8.41$ & $45.65$ & $68.46$ & $2.33$  \\
\textbf{Prefix Tuning (large)} & $1.5\mathbf{M}$ & $65.50$ & $8.45$ & $43.97$ & $67.32$ & $2.23$  \\
\textbf{IA3 (large)} & $0.32\mathbf{M}$ & $65.08$ & $8.5$ & $42.72$ & $66.80$ & $2.15$  \\
\rowcolor{gray!30}
\textbf{RED (large)} & $0.09\mathbf{M}$ & $65.77$ & $8.42$ & $46.12$ & $69.03$ & $2.36$  \\
\Xhline{1pt}
\end{tabular}
\caption{Performance comparison of GPT-2 medium and large models fine-tuned by RED and other PEFT baseline methods on the E2E NLG Challenge. 
The standard deviations of results across different methods are given in Table \ref{tab:standaerd_appendix_gpt2}.}
\label{tab:gpt2}
\vspace{-5mm}
\end{table*}

\subsection{Representation Editing}

Previous PEFT methods refine pre-trained models by updating their weights in a parameter-efficient manner, typically within a low-dimensional space.
To approximate the effect of full-parameter tuning, they are required to choose the values of hyperparameters properly for one downstream task or a set of tasks. 
However, it could be troublesome to choose a suitable value for a hyperparameter, such as the ranks of weight matrices in Equations \eqref{eq:lora} and \eqref{eq:adapter} for LoRA and Adapter respectively.
Enhancing their modeling capacities by using a higher rank $r$ could demand too much computation resources and tend to overfit, while aggressively setting $r$ smaller may degrade model performance and lead to from-scratch re-training \cite{ding2023sparse}.

Our hypothesis posits that by altering the internal representations of neural models rather than their connected weights, the fine-tuning process could be significantly more efficient. 
This is because that direct manipulation of the representations necessitates $O(n)$ parameters, whereas adjusting their weights theoretically demands $O(n^2)$ parameters, with $n$ denoting the dimensionality of hidden representations. \citet{Turner2023ActivationAS} have demonstrated that the behaviors of neural models can be influenced by adding a ``steer vector'' to each hidden layer during inference. 
We postulate that such steer vectors can be learned during the fine-tuning phase. 
Inspired by this concept and the emerging field of representation engineering \citet{Zou2023RepresentationEA}, we introduce a novel PEFT method that fine-tunes the model by directly altering the representation with two learnable vectors, as depicted in Figure \ref{fig:figure_01} (b).

Specifically, we first incorporate a learnable scaling vector $l_\text{scaling}\in \mathbb{R}^{d}$ and apply it to perform the Hadamard product with a hidden representation ${h}_{1}$ by scaling the feature of each dimension within ${h}_{1}$ via element-wise multiplication. 
Additionally, we introduce another learnable bias vector $l_\text{bias} \in \mathbb{R}^{d}$ that is subsequently added to the scaled vector. This process can be formalized as follows:
\begin{equation} \label{eq:red}
 h_{2} = l_\text{scaling} \odot  h_{1} + l_\text{bias}
\end{equation}
\noindent where $\odot$ represents element-wise multiplication, also known as the Hadamard product. Here, $h_{1} \in \mathbb{R}^{d}$ denotes a hidden representation generated by a certain layer, and $h_{2} \in \mathbb{R}^{d}$ represents the resultant edited representation.
During implementation, we initialize the scaling vectors $l_\text{scaling}$ as unit vectors (i.e., with all elements set to $1$) and the bias vectors $l_\text{bias}$ as zero vectors.
This initialization approach ensures that the introduction of these ``edit vectors'' does not initially alter the hidden representations of a neural model.


\vspace{-1mm}
\section{Experiments}
\vspace{-2mm}

Extensive experimentation was conducted to assess the efficacy of our Representation Editing (RED) method across a diverse array of natural language understanding and generation tasks, employing a range of foundational models spanning different scales. 
These models include RoBERTa \cite{Liu2019RoBERTaAR}, T5 \cite{raffel2020exploring}, GPT-2 \cite{Radford2019LanguageMA}, and LLaMA-2 \cite{Touvron2023Llama2O}. 
Specifically, we evaluated RED and the baseline methods on the GLUE benchmark \cite{Wang2018GLUEAM} with RoBERTa and T5, as previously conducted in \cite{hu2021lora} and \citet{Asai2022ATTEMPTPM}.
To ensure consistency with previous studies, we adhered to the experimental setup outlined in \cite{li2021prefix} and \cite{hu2021lora} for comparative analysis with GPT-2.
Moreover, we conducted instruction tuning experiments on LLaMA-2 using the UltraFeedback dataset \cite{Cui2023UltraFeedbackBL} to further assess the applicability of our proposed RED on generative large language models. For further details on the datasets and evaluation metrics used, please refer to Appendix \ref{appendix:dataset}.

\begin{table*}[ht] 
\small
\centering
\setlength{\tabcolsep}{5pt}
\begin{tabular}{l|r|cccccccc|c}
\Xhline{1pt}
\textbf{Method} & \textbf{\# Param} & \textbf{MNLI} & \textbf{SST-2} & \textbf{MRPC} & \textbf{CoLA} & \textbf{QNLI} & \textbf{QQP} & \textbf{RTE} & \textbf{STS-B} & \textbf{Avg.} \\
\Xhline{1pt}
\textbf{FT (base)*} & $220\mathbf{M}$ & $86.8$ & $94.6$ & $90.2$ & $61.8$ & $93.0$ & $91.6$ & $71.9$ & $89.7$ & $84.9$ \\
\textbf{Adapter (base)*} & $1.9\mathbf{M}$ & $86.5$ & $93.8$ & $85.3$ & $64.0$ & $93.2$ & $90.2$ & $71.9$ & $90.7$ & $84.5$ \\
\textbf{AdapterDrop (base)*} & $1.1\mathbf{M}$ & $86.3$ & $93.6$ & $86.3$ & $62.7$ & $93.2$ & $90.2$ & $71.2$ & $91.4$ & $84.4$ \\
\textbf{BitFit (base)*} & $0.3\mathbf{M}$ & $85.3$ & $94.2$ & $86.8$ & $58.2$ & $93.0$ & $90.1$ & $67.6$ & $90.9$ & $83.3$ \\
\textbf{PT (base)*} & $0.08\mathbf{M}$ & $81.3$ & $90.9$ & $68.1$ & $10.6$ & $92.8$ & $89.7$ & $54.7$ & $89.5$ & $72.2$ \\
\rowcolor{gray!30}
\textbf{RED (base)} & $0.04\mathbf{M}$ & $85.9$ & $93.0$ & $91.7$ & $61.1$ & $91.2$ & $89.2$ & $72.7$ & $88.2$ & $84.1$ \\
\Xhline{1pt}
\end{tabular}
\caption{Performance comparison of T5 base fine-tuned by RED and other PEFT baseline methods on the GLUE benchmark. 
Results marked with an asterisk (\textbf{*}) are excerpted from published literature.
}
\vspace{-3mm}
\label{tab:t5}
\end{table*}

\subsection{Baselines}
We ensured that different PEFT methods were compared systematically in a more fair setting. Therefore, we strictly followed the well-established training protocol, evaluating on the validation set after each epoch training, and selecting the checkpoint with the best performance on the validation set as the final model for testing on the test set.
The following baseline methods were used for comparison with the proposed RED:

\begin{itemize}
\setlength{\itemsep}{0pt}
\setlength{\parsep}{0pt}
\setlength{\parskip}{0pt}
\item \textbf{Fine-Tuning (FT)} trains models by updating all their parameters. 
A variant of FT was proposed by \citet{Lee2019WhatWE}, which selectively updates certain layers while freezing others. We incorporate a baseline established in prior research by \citet{li2021prefix} with GPT-2, which specifically adapts only the final two layers, denoted as $\mathbf{FT^{top2}}$.

\item \textbf{Bias-terms Fine-tuning (BitFit)} involves the selective freezing of a majority of the transformer parameters, with the training process focused exclusively on the bias-terms \cite{BenZaken2021BitFitSP}.

\item \textbf{IA3} introduces scaling operations to modify the key and value vectors of the multi-head attentions and those generated by feed-forward networks (FFNs) \cite{Liu2022FewShotPF}.

\item \textbf{Adapter} introduces a learnable, lightweight module situated between two sub-layers of the transformer. 
During the forward pass, inputs are processed in sequence by the sub-layers of the foundation models and the adapters, yielding the final output. However, during the backpropagation phase, only these adapters receive gradients for parameter updates, while the remaining parameters of the model are kept fixed and unaltered \cite{houlsby2019parameter}.

\item \textbf{Adapter-FFN} is a variant of Adapter method proposed by \citet{Pfeiffer2020AdapterFusionNT}.
Contrasting with the original Adapter, which necessitates the insertion of the learnable module between all sub-layers, Adapter-FFN only requires the application of an adapter following each Feed-Forward Network (FFN) sub-layer.

\item \textbf{AdapterDrop} is another variant of Adapter proposed by \citet{Rckl2020AdapterDropOT}, which incorporates a strategy of omitting certain adapter layers, thereby enhancing overall efficiency.

\item \textbf{Low-Rank Adaption (LoRA)} employs a low-rank decomposition on the matrix $\Delta \mathbf{W}$, thereby modeling weight updates as the product of two low-rank matrices.
These two learnable matrices are aligned in parallel with the corresponding matrices in pre-trained models. They process inputs in parallel and combine their results to generate the final outputs in each transformer block \cite{hu2021lora}.

\item \textbf{Prompt Tuning (PT)} prefixes a set of continuous vectors at the embedding layer, which are subjected to learning during the fine-tuning phase \cite{Lester2021ThePO}.

\item \textbf{Prefix Tuning} is a generalized version of prompt tuning (PT), which incorporates learnable continuous vectors at every hidden state. 
These continuous vectors also contribute to the computation of attention, serving as both key and value vectors \cite{Li2021PrefixTuningOC}.
\end{itemize}

\subsection{Results with RoBERTa}

We conducted fine-tuning experiments on both the RoBERTa base ($125$M) and large ($355$M) models by using RED and its competing baselines. 
Subsequently, we evaluated these fine-tuned models on the widely-adopted GLUE benchmark, renowned for its comprehensive evaluation of natural language understanding capabilities.
The pretrained RoBERTa models were sourced from the HuggingFace Transformers Library \cite{Wolf2019HuggingFacesTS}.



We noticed that the previous evaluation settings of PEFT methods encountered two issues.
Firstly, for datasets such as MRPC, RTE, and STS-B, they trained the models on the MNLI dataset first and selecting the best performance checkpoint on the MNLI dataset as initialization for transfer learning to improve the performance of the model trained on these datasets \cite{Liu2019RoBERTaAR,hu2021lora}. However, such a pipeline adds complexity that can be challenging for other researchers to replicate.
Secondly, there is no split between the validation set and the test set. Instead, after each epoch training is completed, evaluation is directly conducted on the test set, and the best test result is selected as the final evaluation result of the model, which violates the established standard that the test set should not influence model selection during the training phase. 
In this study, we ensured a more systematic comparison of different PEFT methods in a fairer setting. To achieve this, we adhered strictly to a well-established training protocol. After each epoch of training, evaluation was conducted on the validation set, and the checkpoint demonstrating the best performance on the validation set was selected as the final model for testing on the test set.
For comprehensive details regarding our reimplementation, please refer to Appendix \ref{appendix:hyperparameter_roberta}.


As presented in Table \ref{tab:roberta}, both the RoBERTa base and large models fine-tuned by RED yielded accuracies comparable to those achieved by other PEFT methods across all tasks within the GLUE benchmark. For instance, RED's performance was only marginally inferior to that of LoRA, differing by a negligible $0.2\%$ with the RoBERTa large and by $0.4\%$ with the RoBERTa base on average, while requiring the tuning of significantly fewer parameters. 
It is noteworthy that RED demonstrated superior performance on tasks such as SST-2, MRPC, CoLA, STS-B, and RTE, all of which had data sizes less than $100$k.
This suggests that RED helps to maintain the generalization capability acquired during the pre-training phase and can deliver enhanced performance when the volume of training data is relatively small. 



Moreover, RED exhibits an unprecedented level of parameter efficiency. 
Although having substantially fewer trainable parameters---approximately $7,200$ times fewer than full parameter fine-tuning and $16$ times fewer than LoRA---it maintains comparable performance. 
This observation suggests that there is still potential for further reduction in the number of trainable parameters and this finding aligns with conclusions drawn by \citet{aghajanyan2020intrinsic} and \citet{Kopiczko2023VeRAVR}.

\subsection{Results with GPT-2}

Beyond natural language understanding tasks, our study expanded to include experiments on natural language generation tasks. 
These experiments were conducted using the GPT-2 medium (355M) and large (774M) models on the E2E NLG Challenge \cite{Novikova2017TheED}. 
The pre-trained GPT-2 models were also obtained from the HuggingFace Transformers Library. 
To ensure a fair comparison, we reproduced other PEFT methods following the settings defined by \citet{li2021prefix} and \citet{hu2021lora}. Please refer to Appendix \ref{appendix:hyperparameter_gpt2} for the comprehensive details of our implementation.



The data presented in Table \ref{tab:gpt2} reveals that RED achieved performance comparable to other PEFT baselines across all metrics in the E2E NLG Challenge. 
This highlights the efficacy of fine-tuning through representation editing, not only for natural language understanding tasks but also for language generation tasks. Remarkably, RED accomplished this performance while still necessitating minimal fine-tuning of parameters.
To compare RED with the most parameter-efficient variants of other PEFT methods, we set the rank of LoRA and Adapter to $1$.
Under these conditions, RED outperformed all such variants while still employing the fewest number of parameters (see Section \ref{sec:rank_1} for details).



\begin{figure}[htp]
\centering
\includegraphics[width=0.40\textwidth]{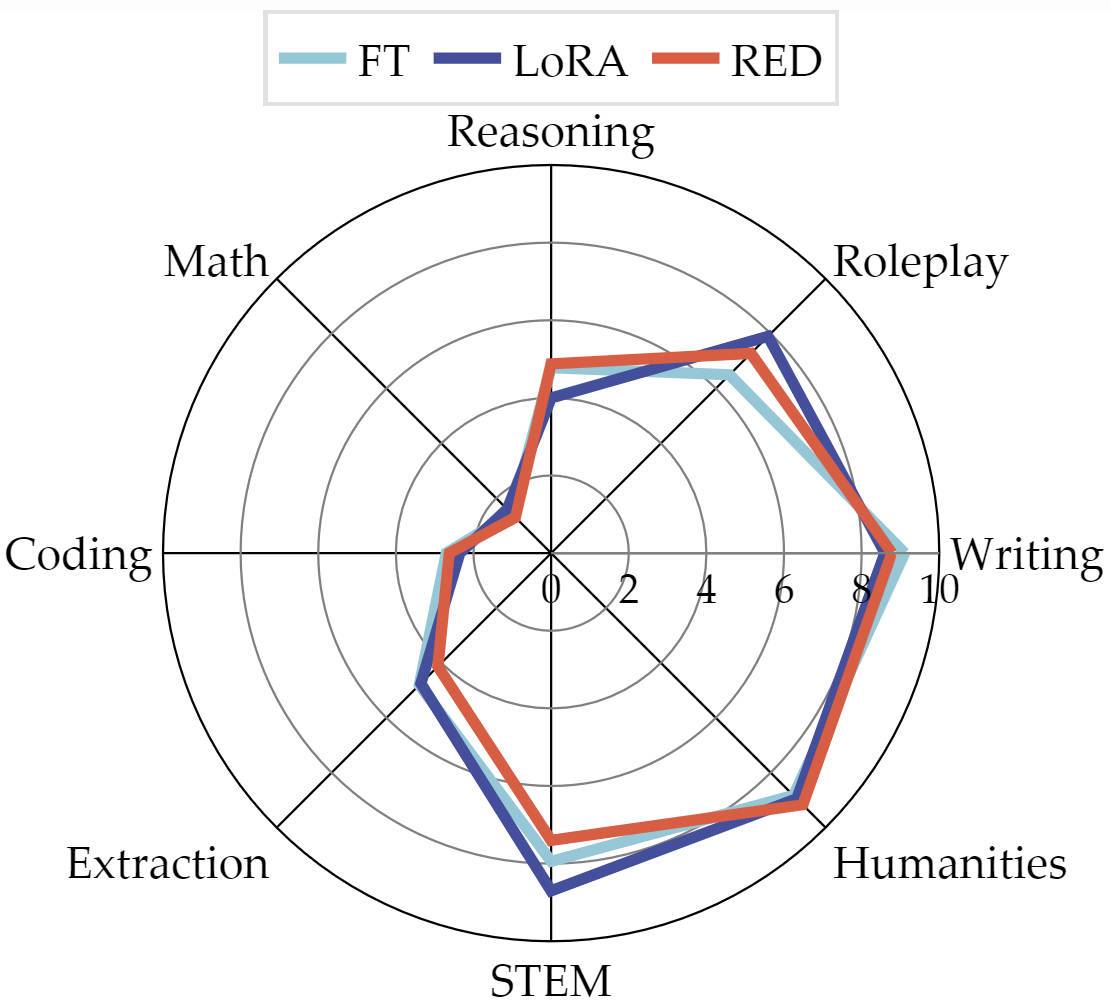}
  \caption{Performance scores achieved by RED and other PEFT methods on the MT-Bench. Refer to Table \ref{tab:mt-details} and Appendix \ref{appendix:llama-2 results} for raw scores and additional details.}
\label{fig:mt-bench}
\vspace{-3mm}
\end{figure}

\begin{table}[ht]
\small
\centering
\setlength{\tabcolsep}{8pt}
\begin{tabular}{l|r|c}
\Xhline{1pt}
\textbf{Method} & \textbf{\# Param} & \textbf{AlpacaEval (win \%)}\\
\hline 
\textbf{FT} & $6739\mathbf{M}$ & $80.93$ \\
\textbf{LoRA} & $8.39\mathbf{M}$ & $81.48$\\
\rowcolor{gray!30}
\textbf{RED} & $0.26\mathbf{M}$ & $81.69$ \\
\Xhline{1pt}
\end{tabular}
\caption{Win rates against reference responses evaluated by GPT-4 on AlpacaEval.
Higher win rates indicate superior generated responses, reflecting better alignment with human preferences. Example responses generated by RED and other PEFT baselines can be found in Figures \ref{Appendix: case1} and \ref{Appendix: case2} as well as in Appendix \ref{appendix:llama-2 results}.
\vspace{-3mm}
}
\label{tab:alpacaeval}
\end{table}

\begin{table*}[t]
\small
\centering
\setlength{\tabcolsep}{8pt}
\begin{tabular}{l|r|ccccc}
\Xhline{1pt}
\textbf{Method} & \textbf{\# Param} & \textbf{BLEU} & \textbf{NIST} & \textbf{MET} & \textbf{ROUGE-L} & \textbf{CIDEr} \\
\Xhline{1pt}
\textbf{Adapter (rank 1)} & $0.25\mathbf{M}$ & $63.76$ & $8.37$ & $42.74$ & $66.70$ & $2.09$  \\
\textbf{Adapter-FFN (rank 1)} & $0.07\mathbf{M}$ & $62.99$ & $8.09$ & $40.88$ & $66.39$ & $1.98$ \\
\textbf{LoRA (rank 1)} & $0.10\mathbf{M}$ & $64.51$ & $8.38$ & $44.78$ & $67.35$ & $2.28$  \\
\rowcolor{gray!30}
\textbf{RED} & $0.05\mathbf{M}$ & $64.86$ & $8.36$ & $44.99$ & $67.62$ & $2.28$  \\
\Xhline{1pt}
\end{tabular}
\caption{Comparison of GPT-2 fine-tuned by RED and other PEFT baselines (rank 1) on the E2E NLG Challenge.}
\label{tab:ablation_rank1}
\vspace{-3mm}
\end{table*}

\subsection{Results with T5}

In order to assess the versatility of RED, we conducted experiments using the encoder-decoder architecture. 
More specifically, we obtained the pre-trained T5-base model ($220$M) from the HuggingFace Transformers Library and evaluated models fine-tuned with RED and other PEFT baselines on the GLUE benchmark. 
As shown in Table \ref{tab:t5}, RED achieved accuracies that were on par with other PEFT methods across all tasks in the GLUE benchmark, indicating its versatility across diverse network architectures.
We excerpted the results from the study of \cite{Asai2022ATTEMPTPM} as they conducted experiments under the settings similar to ours. 
The inclusion of their results does not undermine the fairness or outcome of the comparison.
Please refer to Appendix \ref{appendix:hyperparameter_t5} for the
implementation details.



It is noteworthy that RED surpassed the prompt tuning (PT) baseline with a significant margin of $11.9\%$ on average. 
Despite PT methods necessitating the fewest parameters during fine-tuning among other PEFT baselines, they still require a modification of parameters that is twice as large as that required by RED.
This finding indicates that allocating a few number of learnable parameters to each layer for representation editing is both more parameter-efficient and effective compared to the PT method, which assigns all learnable parameters to the model's embedding layer.




\subsection{Results with LLaMA-2}

We are interested in exploring whether RED can be effectively scaled to large neural networks containing billions of parameters. 
Therefore, we evaluated RED and compared it against full parameter fine-tuning and LoRA using LLaMA-2 up to $7$ billion parameters. Our assessments were conducted across three widely-used benchmarks: Open LLM Leaderboard \cite{open-llm-leaderboard}, AlpacaEval \cite{li2023alpacaeval}, and MT-Bench \cite{Zheng2023JudgingLW}.
The implementation details of this experiment can be found in Appendix \ref{appendix:hyperparameter_llama}.


Table \ref{tab:alpacaeval} presents the win rates obtained by different fine-tuning methods, compared against the reference responses generated by text-davinci-003 on the AlpacaEval. 
Remarkably, RED yielded the highest win rate, even though its number of trainable parameters was $25,700$ times less than that of full parameter fine-tuning and $32$ times less than that of LoRA. 
This suggests that the RED method, which directly edits hidden representations during the fine-tuning phase, continues to be effective for large language models, and the trained models are capable of generating high-quality responses.



Moreover, Figure \ref{fig:mt-bench} illustrates the performance scores achieved by three training methods on the 1-turn question-answer dataset from MT-Bench. The overall performance of RED is on par with other baseline methods, and it notably excels in assessing the capabilities of humanities and reasoning. As presented in Table \ref{tab:openllm}, RED also delivers competitive results on six distinct datasets from the Open LLM Leaderboard.


\begin{table}[htbp]
\small
\centering
\setlength{\tabcolsep}{6pt}
\begin{tabular}{l|cccc}
\Xhline{1pt} 
\textbf{Method} & \textbf{MRPC} & \textbf{CoLA} & \textbf{QQP} & \textbf{Avg.} \\
\Xhline{1pt} 
\rowcolor{gray!30}
\textbf{RED} & $90.3$ & $68.1$ & $88.7$ & $82.4$ \\
${\textbf{w/o scaling}}$ & $89.8$ & $65.9$ & $87.6$ & $81.1$ \\
${\textbf{w/o bias}}$ & $75.8$ & $46.9$ & $87.2$ & $70.0$ \\
 \Xhline{1pt} 
\end{tabular}
\caption{
Results of the ablation studies on three different datasets from the GLUE benchmark. The term ``w/o scaling'' denotes the removal of scaling operations, with representation editing performed only by bias vectors. In contrast, ``w/o bias'' indicates the omission of bias vectors, with representation editing conducted exclusively through the use of scaling vectors.
}
\label{tab:bias_scaling}
\vspace{-5mm}
\end{table}

\section{Ablation Study}
\vspace{-2mm}

In this section, we conducted a series of ablation studies to examine the impact of different editing operations used in RED and to explore potential application points within transformer-based networks. We also evaluated the parameter efficiency of RED by comparing it with the most parameter-efficient variants of other PEFT methods.


\begin{table*}[ht] 
\small
\centering
\setlength{\tabcolsep}{5pt}
\begin{tabular}{l|r|cccccccc|c}
\Xhline{1pt}
\textbf{Method} & \textbf{\# Param} & \textbf{MNLI} & \textbf{SST-2} & \textbf{MRPC} & \textbf{CoLA} & \textbf{QNLI} & \textbf{QQP} & \textbf{RTE} & \textbf{STS-B} & \textbf{Avg.} \\
\Xhline{1pt}
\textbf{LoRA (base)} & $0.29\mathbf{M}$ & $86.6$ & $93.9$ & $88.7$ & $59.7$ & $92.6$ & $90.4$ & $75.3$ & $90.3$ & $84.7$ \\
\textbf{RED (base)} & $0.02 \mathbf{M}$ & $83.9$ & $93.9$ & $89.2$ & $61.0$ & $90.7$ & $87.2$ & $78.0$ & $90.4$ & $84.3$ \\
\rowcolor{gray!30}
\textbf{RED+ (base)} & $0.09 \mathbf{M}$ & $85.9$ & $93.2$ & $88.9$ & $62.4$ & $90.9$ & $89.1$ & $78.6$ & $90.9$ & $85.0$ \\
\hline
\textbf{LoRA (large)} & $0.79\mathbf{M}$ & $90.2$ & $96.0$ & $89.8$ & $65.5$ & $94.7$ & $90.7$ & $86.3$ & $91.7$ & $88.1$ \\
\textbf{RED (large)} & $0.05\mathbf{M}$ & $89.5$ & $96.0$ & $90.3$ & $68.1$ & $93.5$ & $88.8$ & $86.2$ & $91.3$ & $87.9$ \\
\rowcolor{gray!30}
\textbf{RED+ (large)} & $0.25\mathbf{M}$ & $90.6$ & $95.6$ & $89.2$ & $69.6$ & $94.0$ & $90.2$ & $85.6$ & $91.4$ & $88.3$ \\
\Xhline{1pt}
\end{tabular}
\vspace{-1mm}
\caption{Performance comparison of RoBERTa base and large models fine-tuned by RED, RED+ and other PEFT baselines on the GLUE benchmark. RED+ denotes an enhanced variant of RED where the representation vectors of $Q$, $K$, and $V$ in attention blocks are also modified via representation editing in addition to FFN sub-layers. 
}
\label{tab:roberta++}
\vspace{-2mm}
\end{table*}

\begin{table*}[ht] 
\small
\centering
\setlength{\tabcolsep}{8pt}
\begin{tabular}{l|r|ccccc}
\Xhline{1pt}
\textbf{Method} & \textbf{\# Param} & \textbf{BLEU} & \textbf{NIST} & \textbf{MET} & \textbf{ROUGE-L} & \textbf{CIDEr} \\
\Xhline{1pt}
\textbf{LoRA (medium)} & $0.79\mathbf{M}$ & $67.43$ & $8.65$ & $46.01$ & $69.64$ & $2.42$  \\
\textbf{RED (medium)} & $0.05\mathbf{M}$ & $64.86$ & $8.36$ & $44.99$ & $67.62$ & $2.28$  \\
\rowcolor{gray!30}
\textbf{RED+ (medium)} & $0.25\mathbf{M}$ & $66.68$ & $8.53$ & $46.28$ & $69.63$ & $2.38$  \\
\hline
\textbf{LoRA (large)} & $1.47\mathbf{M}$ & $68.24$ & $8.76$ & $46.23$ & $69.92$ & $2.42$  \\
\textbf{RED (large)} & $0.09\mathbf{M}$ & $65.77$ & $8.42$ & $46.12$ & $69.03$ & $2.36$  \\
\rowcolor{gray!30}
\textbf{RED+ (large)} & $0.46\mathbf{M}$ & $68.31$ & $8.78$ & $46.12$ & $69.80$ & $2.41$  \\
\Xhline{1pt}
\end{tabular}
\vspace{-1mm}
\caption{Performance comparison of GPT-2 medium and large models fine-tuned by RED, RED+ and other PEFT baselines on the E2E NLG Challenge. RED+ denotes an enhanced variant of RED where the representation vectors of $Q$, $K$, and $V$ in attention blocks are also modified via representation editing in addition to FFN sub-layers. 
}
\label{tab:gpt2++}
\vspace{-5mm}
\end{table*}

\subsection{Impact of Different Editing Operators}

We introduce two operators to edit the hidden representations of neural network models: one scales a hidden representation using a learnable vector, and the other adds a bias vector to the hidden representation. To assess the contribution of each operator, we performed ablation studies by removing one operator at a time and examining the impact on the models' performance.


From Table \ref{tab:bias_scaling}, it is evident that both editing operators are crucial for enhancing model performance. 
Analyzing results from three different datasets from the GLUE benchmark, we found that the scaling operator boosted accuracy by an average of $1.3\%$, while the biasing operator contributed to a significant $12.4\%$ increase in accuracy. 
Compared to the scaling operator, the addition of a learnable bias vector to the original representation plays a more substantial role in the fine-tuning of neural models.



\subsection{Impact of Editing Positions}

The motivation of this study is to explore the feasibility of fine-tuning large language models with as few parameters as possible. Therefore, in our current implementation, we only consider modifying the representations produced by FFN sub-layers. However, it is still possible to further improve the performance of the model by increasing the number of trainable parameters if the representation vectors of $Q$ (query), $K$ (key), and $V$ (value) in attention blocks are allowed to be modified via representation editing. To demonstrate this possibility, we conducted additional experiments in which the representations of $Q$, $K$, and $V$ were also modified. 


The experimental results reported in Table \ref{tab:roberta++} and Table \ref{tab:gpt2++} show that the performance of RED has been improved on all these datasets but with slightly more trainable parameters (increasing from $0.07$M to $0.37$M). It suggests that it is feasible to increase the number of trainable parameters to further boost the performance by editing additional feature representations.
This strategy is particularly effective for datasets with a larger number of training instances, such as MNLI, QNLI, QQP, and E2E while the performance did not exhibit significant improvement on datasets with fewer training examples.

\subsection{Parameter Efficiency and Efficacy}
\label{sec:rank_1}

When using Adaptor or LoRA for fine-tuning, the ranks of their decomposition matrices  were typically set to $8$ or $16$.
These ranks can be reduced to decrease the number of parameters used during fine-tuning. 
To compare RED with the most parameter-efficient versions of Adpator or LoRA, we adjusted the rank of their decomposition matrices to $1$. 
We then made a comparative analysis with GPT-2 medium on the E2E NLG Challenge.

Table \ref{tab:ablation_rank1} shows that RED outperformed all the baselines on four out of five tasks, while it slightly underperformed LoRA by a negligible difference of $0.02$ on the NIST dataset. 
It is worthy noting that RED accomplished this with  the minimal number of parameters, demonstrating that RED is not only parameter-efficient but also parameter-effective.



\section{Conclusion}


In this study, we proposed a novel perspective on fine-tuning by editing model representations rather than adjusting the model's weights.
Through extensive experimentation across various NLP tasks and neural models of varying structures and scales, we demonstrated that RED can deliver performance comparable or superior to existing PEFT methods while employing a minimal number of trainable parameters.
We are hopeful that this study could inspire novel methods for fine-tuning LLMs, thereby contributing to the advancement of both effective and computationally efficient PEFT techniques.



\section*{Limitations}

We have demonstrated the efficacy of a new promising PEFT approach for fine-tuning models through direct manipulation of representations across various NLP tasks with networks of varying structures and scales. 
It would be intriguing to extend this method to other modalities, such as image, speech, and video.
Recent advancements in representation engineering have indicated that only a very few examples are sufficient for achieving precise control over the model's output through representation manipulations. 
We plan to apply the proposed RED to few-shot scenarios in the future.


\section*{Reproducibility Statement}

We have made our code publicly accessible via a repository on GitHub (github.com/mlwu22/RED). To further ensure replicability, we had a colleague unfamiliar with our method install and test RED. The experiment yielded results nearly identical to ours, which strengthens our confidence that other researchers will successfully execute our code and reproduce our findings.


\bibliography{custom}

\clearpage

\appendix

\section{Datasets}
\label{appendix:dataset}

\subsection{GLUE Benchmark}

The General Language Understanding Evaluation (GLUE) benchmark comprises a variety of datasets, including CoLA \cite{Warstadt2018NeuralNA}, SST-2 \cite{Socher2013RecursiveDM}, MRPC \cite{Dolan2005AutomaticallyCA}, QQP \cite{Wang2018GLUEAM}, STS-B \cite{Cer2017SemEval2017T1}, MNLI \cite{Williams2017ABC}, QNLI \cite{Demszky2018TransformingQA}, and RTE \cite{BarHaim2006TheSP}. 
These datasets have been widely-used to measure the performance of language understanding models.
The GLUE benchmark was sourced from Huggingface Datasets \cite{Lhoest2021DatasetsAC}, and all datasets within this benchmark were employed for our evaluation.


Following \cite{ding2023sparse} and \cite{hu2021lora}, we evaluated models trained with different PEFT methods on the validation dataset.
To ensure a fair and systematic comparison of different PEFT methods, we randomly sampled $1,000$ instances from the validation set to form a new validation set if the validation set contains more than $2,000$ instances (otherwise, we randomly selected half of the instances for the new validation set), and the remaining instances were used as the test set.
This allowed us to adhere rigorously to the established training protocol, which involves evaluating on the new validation set after each epoch of training and selecting the checkpoint that yields the best performance on the validation set as the final model for testing on the test set.
The sizes of the training, validation, and test sets as well as their corresponding metrics for all datasets are given in Table \ref{tab:dataset_appendix}.


For all the experiments with RoBERTa, we ran experiment $5$ times for each PEFT method using distinct random seeds to initialize the trainable weights and reported the average results.
During the evaluation of the RTE and CoLA datasets, some researchers reported that specific random seeds could potentially lead to anomalous experimental outcomes\footnote{Refer to \url{github.com/microsoft/LoRA/issues}.}. 
Therefore, we randomly chose other five different random seeds (i.e., $42$, $43$, $44$, $45$, and $46$) to conduct the experiments.


\begin{table}[ht] 
\small
\centering
\setlength{\tabcolsep}{5pt}
\begin{tabular}{l|rccc}
\Xhline{1pt}
\textbf{Dataset} & \textbf{\#Train} & \textbf{\#Validation} & \textbf{\#Test} & \textbf{Metric} \\
\Xhline{1pt}
\textbf{CoLA} & $8.5$K & $522$ & $521$ & MCC \\
\textbf{SST-2} & $67$K & $436$ & $436$ & ACC \\
\textbf{MRPC} & $3.7$K & $204$ & $204$ & ACC \\
\textbf{QQP} & $364$K & $1$K & $39$K & ACC \\
\textbf{STS-B} & $5.7$K & $750$ & $750$ & CORR \\
\textbf{MNLI} & $393$K & $1$K & $8$K & ACC \\
\textbf{QNLI} & $105$K & $1$K & $4.5$K & ACC \\
\textbf{RTE} & $2.5$K & $139$ & $138$ & ACC \\
\Xhline{1pt}
\end{tabular}
\caption{The sizes of the training, validation, and test sets, along with their corresponding metrics for all datasets in the GLUE benchmark.
``MCC'', ``ACC'' and ``CORR'' denote Matthews correlation coefficient, accuracy, and Pearson correlation coefficient respectively.}
\label{tab:dataset_appendix}
\vspace{-3mm}
\end{table}

\subsection{E2E NLG Challenge}

The E2E NLG Challenge was first introduced by \citet{Novikova2017TheED} to train and evaluate end-to-end, data-driven natural language generation models and systems. 
All datasets in the E2E NLG Challenge were sourced from Huggingface Datasets. 
This benchmark includes $42.1$K training instances, $4.67$K validation instances, and $4.69$K testing instances. 
Following previous studies, we used the official evaluation script to compute BLEU \cite{Papineni2002BleuAM}, NIST \cite{Belz2006ComparingAA}, METEOR \cite{Banerjee2005METEORAA}, ROUGE-L \cite{Lin2004ROUGEAP} and CIDEr \cite{Vedantam2014CIDErCI} scores. 
For evaluations using GPT-2,  we ran experiment for $3$ times using distinct random seeds and reported the average results. 
The random seeds used for these experiments were $42$, $43$, and $44$.

\subsection{UltraFeedback}

UltraFeedback \cite{Cui2023UltraFeedbackBL} comprises $64,000$ prompts, each of them accompanied by four LLM responses evaluated by GPT-4 based on criteria such as instruction-following, honesty, and helpfulness.
To construct the training dataset from UltraFeedback, we selected the responses with the highest mean scores\footnote{The constructed dataset can be accessed at: \\
\url{huggingface.co/datasets/argilla/ \\ ultrafeedback-binarized-preferences-cleaned}.}.

\subsection{Open LLM Leaderboard}

The Open LLM Leaderboard includes six datasets that assess various capabilities such as science questions, commonsense inference, multitask accuracy, mathematical reasoning, and the truthfulness of generated answers. 
It consists of ARC \cite{Mihaylov2018CanAS}, HellaSwag \cite{Zellers2019HellaSwagCA}, WinoGrande \cite{Sakaguchi2019AnAW}, MMLU \cite{Hendrycks2020MeasuringMM}, TruthfulQA \cite{Lin2021TruthfulQAMH}, and GSM8K \cite{Hendrycks2021MeasuringMP}. 
We used the Eleuther AI Language Model Evaluation Harness library \cite{eval-harness} to evaluate language models trained using various methods. Table \ref{tab: Open-LLM-Leaderboard} details the leaderboard evaluation configuration and the experimental settings applied in this study.


\subsection{AlpacaEval}

AlpacaEval is an automated evaluation benchmark for LLMs, employing GPT-4 \cite{gpt4} as an annotator to compare the responses generated by the models against reference answers produced by text-davinci-003. The benchmark consists of 805 samples focused on simple instruction-following tasks. Previous research has shown a high correlation between GPT-4's annotations and human evaluator assessments \cite{li2023alpacaeval}.


\subsection{MT-Bench}

MT-Bench \cite{Zheng2023JudgingLW} comprises $80$ challenging questions, each structured as a two-turn dialogue. GPT-4 was also used to evaluate the responses generated by the models, assigning a score from $1$ to $10$ for each turn.


\section{Hyperparameters}
\label{appendix:hyperparameter}

\subsection{Experiments with RoBERTa}
\label{appendix:hyperparameter_roberta}

We trained the models using AdamW with a linear learning rate decay schedule. To ensure a fair comparison, we maintained the same sequence length across all baseline PEFT methods. 
The hyperparameters used in our experiments for RoBERTa base are detailed in Table \ref{tab:hyper_appendix_roberta_base}, and for RoBERTa large in Table \ref{tab:hyper_appendix_roberta_large}.





\subsection{Experiments with GPT-2}
\label{appendix:hyperparameter_gpt2}

We trained the models using AdamW with a linear learning rate decay schedule. 
To ensure a fair comparison, we maintained the same sequence length across all baseline PEFT methods. 
We used the Huggingface PEFT package to replicate Prefix Tuning and LoRA, and the OpenDelta package to replicate Adapter and Adapter-FFN. 
The hyperparameters used in our experiments for GPT-2 medium and GPT-2 large are detailed in Table \ref{tab:hyper_appendix_gpt2}s.



\subsection{Experiments with T5}
\label{appendix:hyperparameter_t5}

We trained the models using AdamW with a linear learning rate decay. 
We conducted experiments on T5 following the settings established by \citet{Asai2022ATTEMPTPM}. 
The hyperparameters used in our experiments for T5 base are detailed in Table \ref{tab:hyper_appendix_t5_base}. 


\subsection{Experiments with LLaMA-2}
\label{appendix:hyperparameter_llama}

We employed AdamW with a cosine learning rate decay schedule during the fine-tuning. To ensure a fair comparison, we maintained the same sequence length across all baseline methods. 
The hyperparameters used for LLaMA-2 are listed in Table \ref{tab:hyper_appendix_Llama2}. 
We evaluated the performance of models on TruthfulQA after each training epoch, and the checkpoint demonstrating the best performance was selected for final testing on other datasets. 
A greedy decoding strategy was employed for generation for all the models. 
To mitigate repetition issues, we applied a repetition penalty of $1.1$ and set the no-repeat $n$-gram size to $5$.

\vspace{-2mm}
\section{Results of Standard Deviations}
\label{appendix:standard deviations}
\vspace{-2mm}

We presented the average results in the main text. 
To provide a quantifiable measure of uncertainty in the measurement or prediction and help other researchers replicate our experiments, we also reported the standard deviations with RoBERTa models on the GLUE benchmark in Table \ref{tab:standaerd_appendix_roberta}, and the standard deviations with GPT-2 on the E2E NLG Challenge in Table \ref{tab:standaerd_appendix_gpt2}.


\vspace{-1mm}
\section{Additional Results with LLaMA-2}
\label{appendix:llama-2 results}
\vspace{-2mm}

\begin{figure*}[t]
    \centering
    \includegraphics[width=0.95\linewidth]{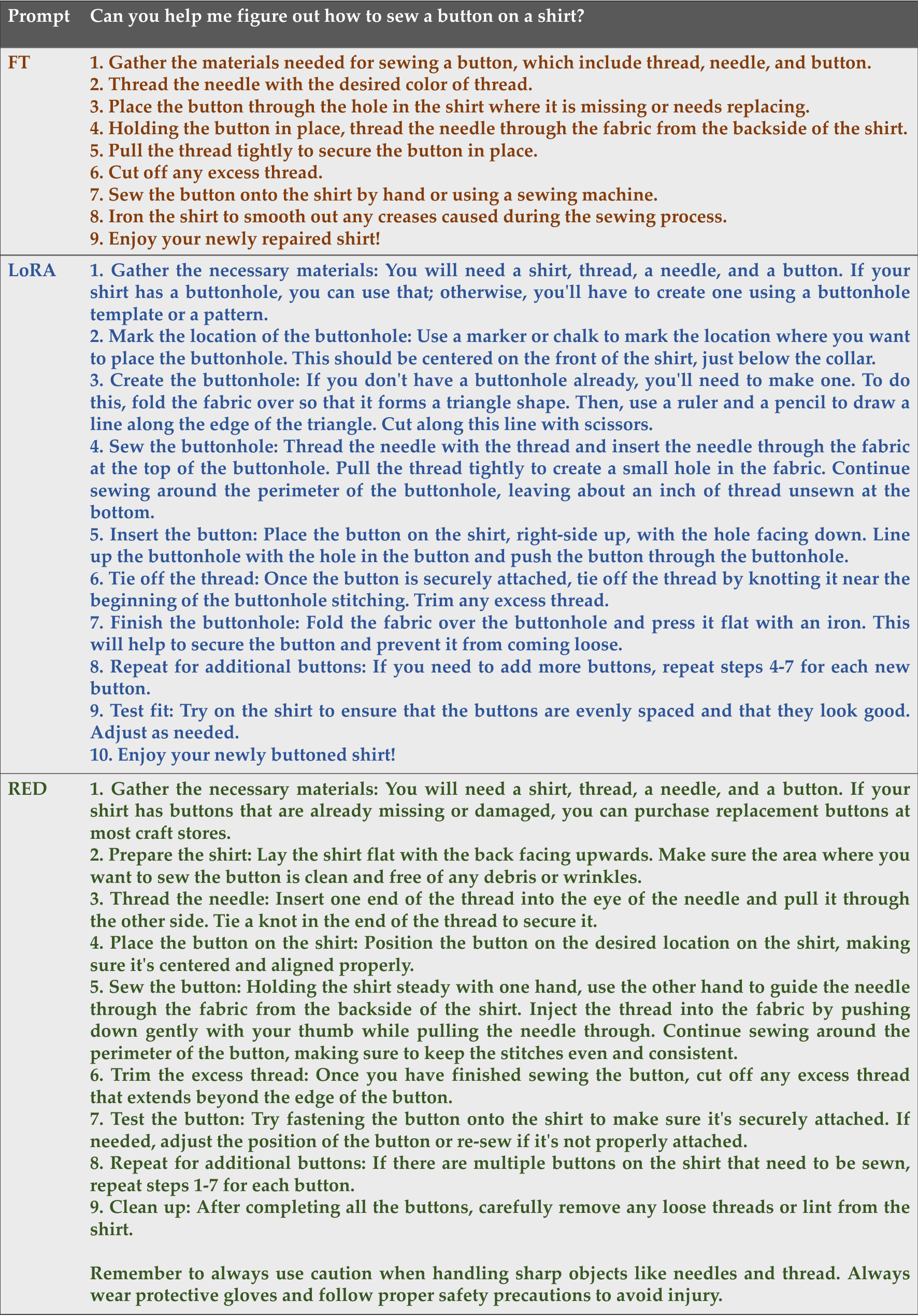}
    \caption{The model fine-tuned with RED generates a thorough, sequential guide that offers accurate details, facilitating comprehension even for novices. This guide encompasses elements such as preparation, threading, positioning, sewing techniques, and post-sewing cleanup, while also providing safety advice and promoting testing for secure attachment. In contrast, the LoRA-trained response inaccurately concentrates on buttonhole creation rather than button sewing, potentially leading to confusion for individuals seeking button attachment guidance. The response generated by the full-parameter trained model presents a simplified summary, but it lacks the in-depth explanation and precision of RED, rendering it less informative for those unfamiliar with the sewing process.}
    \label{Appendix: case1}
\end{figure*}

\begin{figure*}[t]
    \centering
    \includegraphics[width=0.95\linewidth]{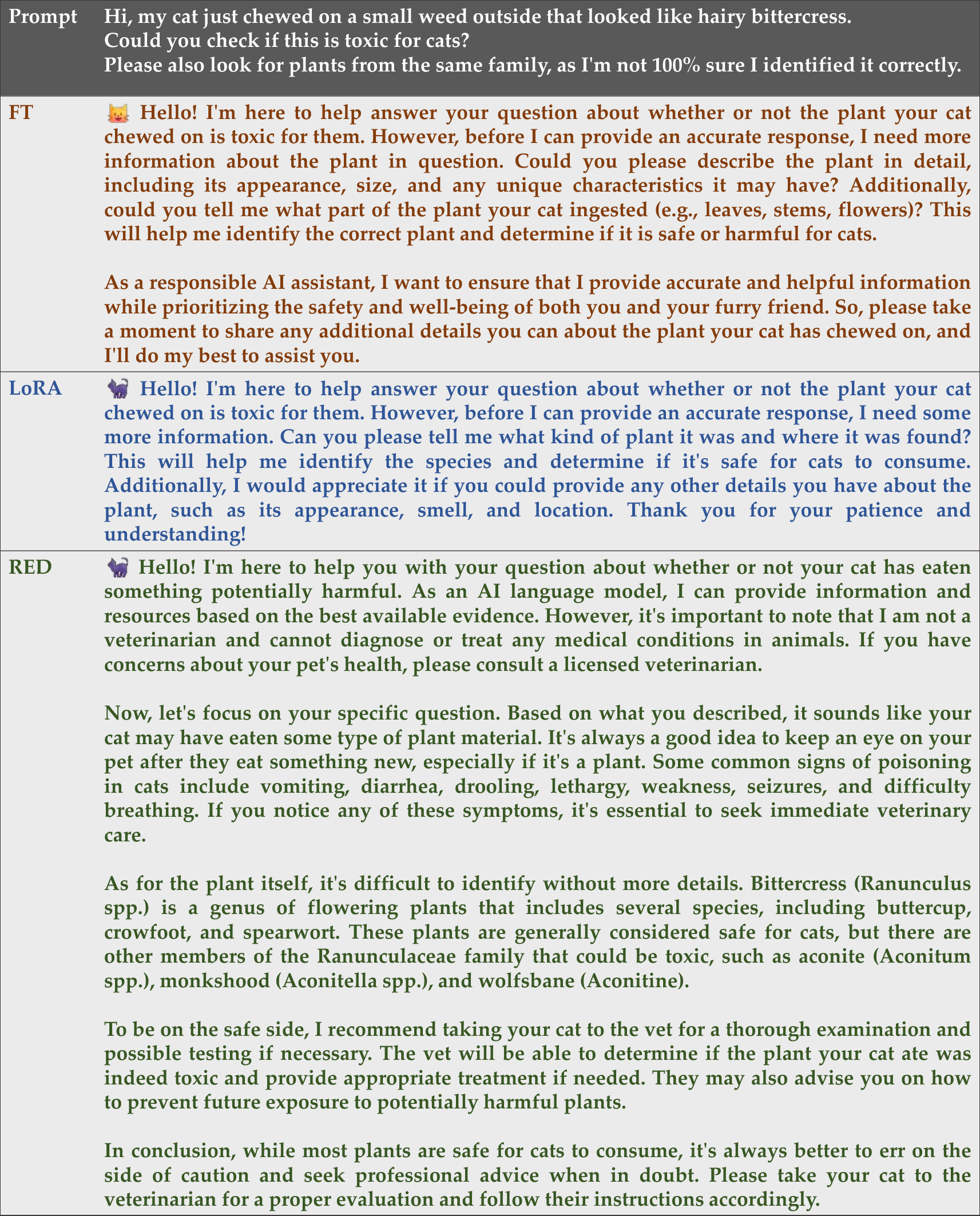}
    \caption{The model fine-tuned using RED generates a comprehensive and proactive strategy, addressing immediate issues, potential symptoms to monitor, and the significance of veterinary consultation. It offers an overarching safety evaluation of the Ranunculaceae family, indicating potentially toxic members and highlighting the necessity for professional assessment. This response strikes a balance between informative content and practical guidance, empowering pet owners to act in their pet's best interests, even in the absence of specific plant identification. In contrast, the responses produced by the models trained with full parameters and LoRA place a greater emphasis on collecting further information before offering advice, which could inadvertently postpone critical care in an emergent situation.}
    \label{Appendix: case2}
\end{figure*}

Example responses generated by models trained with RAD, full-parameter fine-tuning (FT) and LoRA are presented in Figures \ref{Appendix: case1} and \ref{Appendix: case2}.
Tables \ref{tab:mt-details} and \ref{tab:openllm} provide the results on the MT-Bench and Open LLM Leaderboard respectively.

\vspace{-2mm}
\begin{table}[ht]
\small
\centering
\setlength{\tabcolsep}{5pt}
\begin{tabular}{l|l|c}
\Xhline{1pt}
\textbf{Method} & \textbf{Hyperparameter} & \textbf{Value} \\
\Xhline{1pt}
\multirow{6}{*}{\textbf{ALL}} & Batch Size         & $128$ \\
&Micro Batch Size   &$2$ \\
&Optimizer &Adamw \\
&LR Scheduler Type & Cosine\\
&Warmup Ratio &$0.1$\\
&Maximum Sequence Length & $768$ \\
\hline
\multirow{2}{*}{\textbf{FT}} & Learning Rate   & $2e-5$\\
& Number of Epochs           & $3$  \\
\hline
\multirow{5}{*}{\textbf{LoRA}} & Learning Rate   & $3e-4$\\
& Number of Epochs           & $3$  \\
&Batch Size         & $128$ \\
&Coefficient $\alpha$ & $16$\\
&Dropout & $0.1$\\
&Rank & $16$\\
\hline
\multirow{2}{*}{\textbf{RED}} & Learning Rate   & $1e-3$\\
& Number of Epochs           & $10$  \\
\Xhline{1pt}
\end{tabular}
\vspace{-1mm}
\caption{Hyperparameters used for training LLaMA-2.}
\label{tab:hyper_appendix_Llama2}
\vspace{-3mm}
\end{table}

\vspace{-2.5mm}
\begin{table}[ht]
\small
\centering
\setlength{\tabcolsep}{8pt}
\begin{tabular}{l|c|c}
\Xhline{1pt}
\textbf{Dataset} & \textbf{\# Few-shot} & \textbf{Metric} \\
\Xhline{1pt}
\textbf{ARC} & $25$ & ACC-NORM \\
\textbf{TruthfulQA} & $0$ & MC2 \\
\textbf{Winogrande} & $5$ & ACC \\
\textbf{GSM8K} & $5$ & ACC \\
\textbf{HellaSwag} & $10$ & ACC-NORM \\
\textbf{MMLU} & $10$ & ACC \\
\Xhline{1pt}
\end{tabular}
\vspace{-2mm}
\caption{Number of few-shot examples and metrics used for evaluation on the datasets in the Open LLM Leaderboard. ``MC2'' denotes the normalized total probability assigned to the set of true answers, and ``ACC-NORM'' denotes the normalized accuracy used in settings where response lengths can vary significantly.}
\label{tab: Open-LLM-Leaderboard} 
\vspace{-3mm}
\end{table}

\begin{table*}[t]
\small
\centering
\setlength{\tabcolsep}{8pt}
\begin{tabular}{l|r|cccccc|c}
\Xhline{1pt} 
\textbf{Method}   & \textbf{\# Param}   & \textbf{ARC}     & \textbf{TruthfulQA} & \textbf{Winogrande}    & \textbf{GSM8k} & \textbf{HellaSwag} & \textbf{MMLU} & \textbf{Average}  \\
\Xhline{1pt}
\textbf{FT}   & $6739\mathbf{M}$ & $73.34$ & $47.07$   & $74.03$ & $16.22$  &  $79.95$  & $46.55$   & $56.19$ \\
\textbf{LoRA}  & $8.39\mathbf{M}$  & $72.07$ & $44.84$   & $75.85$ & $15.01$  &  $78.60$  & $44.42$  & $55.13$ \\
\rowcolor{gray!30}
\textbf{RED} & $0.26\mathbf{M}$  & $72.04$ & $47.59$   & $72.85$ & $11.90$  &  $77.86$  & $42.27$  & $54.09$ \\
\Xhline{1pt} 
\end{tabular}
\caption{Results achieved by RED, LoRA and full-parameter fine-tuning (FT) on the Open LLM Leaderboard.}
\label{tab:openllm}   
\end{table*}

\begin{table*}[ht] 
\small
\centering
\setlength{\tabcolsep}{3pt}
\begin{tabular}{l|l|cccccccc}
\Xhline{1pt} 
\textbf{Method}  & \textbf{Dataset} & \textbf{MNLI} & \textbf{SST-2} & \textbf{MRPC} & \textbf{CoLA} & \textbf{QNLI} & \textbf{QQP} & \textbf{RTE} & \textbf{STS-B} \\
\Xhline{1pt}
\multirow{2}{*}{\textbf{ALL}} & {Optimizer} &  \multicolumn{8}{c}{AdamW}\\
& LR Schedule &  \multicolumn{8}{c}{Linear}\\
\hline
\multirow{5}{*}{\textbf{FT}} & Batch Size & $32$ & $32$ & $32$ & $32$ & $32$ & $32$ & $32$ & $32$ \\
& Number of Epochs & $20$ & $40$ & $40$ & $40$ & $20$ & $20$ & $40$ & $40$ \\
&  Learning Rate & $1e-05$ & $1e-05$ & $1e-05$ & $1e-05$ & $1e-05$ & $1e-05$ & $1e-05$ & $1e-05$ \\
& Warmup Ratio & $6e-02$ & $6e-02$ & $6e-02$ & $6e-02$ & $6e-02$ & $6e-02$ & $6e-02$ & $6e-02$ \\ 
&  Weight Decay & \multicolumn{8}{c}{$1e-04$} \\
&  Maximum Sequence Length & \multicolumn{8}{c}{$256$} \\
\hline
\multirow{5}{*}{\textbf{LoRA}} & Batch Size & $32$ & $32$ & $32$ & $32$ & $32$ & $32$ & $32$ & $32$ \\
& Number of Epochs & $20$ & $40$ & $40$ & $40$ & $20$ & $20$ & $40$ & $40$ \\
&  Learning Rate & $5e-04$ & $5e-04$ & $4e-04$ & $4e-04$ & $4e-04$ & $5e-04$ & $5e-04$ & $4e-04$ \\
& Warmup Ratio & $6e-02$ & $6e-02$ & $6e-02$ & $6e-02$ & $6e-02$ & $6e-02$ & $6e-02$ & $6e-02$ \\ 
& Rank & \multicolumn{8}{c}{$8$} \\
& Coefficient $\alpha$ & \multicolumn{8}{c}{$8$} \\
&  Maximum Sequence Length & \multicolumn{8}{c}{$256$} \\
\hline
\multirow{5}{*}{\textbf{Adapter}} & Batch Size & $32$ & $32$ & $32$ & $32$ & $32$ & $32$ & $32$ & $32$ \\
& Number of Epochs & $20$ & $40$ & $40$ & $40$ & $20$ & $20$ & $40$ & $40$ \\
&  Learning Rate & $1e-04$ & $1e-04$ & $1e-04$ & $1e-04$ & $1e-04$ & $1e-04$ & $1e-04$ & $1e-04$ \\
& Warmup Ratio & $6e-02$ & $6e-02$ & $6e-02$ & $6e-02$ & $6e-02$ & $6e-02$ & $6e-02$ & $6e-02$ \\ 
& Rank & \multicolumn{8}{c}{$8$} \\
&  Maximum Sequence Length & \multicolumn{8}{c}{$256$} \\
\hline
\multirow{5}{*}{\textbf{Adapter-FFN}} & Batch Size & $32$ & $32$ & $32$ & $32$ & $32$ & $32$ & $32$ & $32$ \\
& Number fo Epochs & $20$ & $40$ & $40$ & $40$ & $20$ & $20$ & $40$ & $40$ \\
 &  Learning Rate & $1e-04$ & $1e-04$ & $1e-04$ & $1e-04$ & $1e-04$ & $1e-04$ & $1e-04$ & $1e-04$ \\
& Warmup Ratio & $6e-02$ & $6e-02$ & $6e-02$ & $6e-02$ & $6e-02$ & $6e-02$ & $6e-02$ & $6e-02$ \\ 
& Rank & \multicolumn{8}{c}{$16$} \\
&  Maximum Sequence Length & \multicolumn{8}{c}{$256$} \\
\hline
\multirow{5}{*}{\textbf{BitFit}} & Batch Size & $32$ & $32$ & $32$ & $32$ & $32$ & $32$ & $32$ & $32$ \\
& Number of Epochs & $20$ & $40$ & $40$ & $40$ & $20$ & $20$ & $40$ & $40$ \\
&  Learning Rate & $1e-04$ & $1e-04$ & $1e-04$ & $1e-04$ & $1e-04$ & $1e-04$ & $1e-04$ & $1e-04$ \\
& Warmup Ratio & $6e-02$ & $6e-02$ & $6e-02$ & $6e-02$ & $6e-02$ & $6e-02$ & $6e-02$ & $6e-02$ \\ 
& Maximum Sequence Length & \multicolumn{8}{c}{$256$} \\
\hline
\multirow{5}{*}{\textbf{RED}} & Batch Size & $32$ & $32$ & $32$ & $32$ & $32$ & $32$ & $32$ & $32$ \\
& Number of Epochs & $20$ & $40$ & $40$ & $40$ & $20$ & $20$ & $40$ & $40$ \\
&  Learning rate & $1e-03$ & $1e-03$ & $5e-03$ & $5e-03$ & $3e-03$ & $1e-03$ & $8e-03$ & $3e-03$ \\
& Warmup Ratio & $6e-02$ & $6e-02$ & $6e-02$ & $6e-02$ & $6e-02$ & $6e-02$ & $6e-02$ & $6e-02$ \\ 
& Maximum Sequence Length & \multicolumn{8}{c}{$256$} \\
\Xhline{1pt}
\end{tabular}
\caption{Hyperparameter values for training RoBERTa base on the GLUE benchmark.}
\label{tab:hyper_appendix_roberta_base}
\end{table*}

\begin{table*}[ht] 
\small
\centering
\setlength{\tabcolsep}{4.5pt}
\begin{tabular}{l|l|cccccccc}
\Xhline{1pt}
\textbf{Method} & \textbf{Dataset} & \textbf{MNLI} & \textbf{SST-2} & \textbf{MRPC} & \textbf{CoLA} & \textbf{QNLI} & \textbf{QQP} & \textbf{RTE} & \textbf{STS-B} \\
\Xhline{1pt} 
\multirow{2}{*}{\textbf{ALL}} & {Optimizer} &  \multicolumn{8}{c}{AdamW}\\
& LR Schedule &  \multicolumn{8}{c}{Linear}\\
\hline
\multirow{5}{*}{\textbf{RED}} & Batch Size & $32$ & $32$ & $32$ & $32$ & $32$ & $32$ & $32$ & $32$ \\
& Number of Epochs & $10$ & $10$ & $20$ & $60$ & $10$ & $10$ & $30$ & $50$ \\
&  Learning rate & $5e-02$ & $3e-02$ & $1e-01$ & $4e-02$ & $2e-02$ & $5e-03$ & $7e-02$ & $4e-02$ \\
& Warmup Ratio & $1e-02$ & $6e-02$ & $1e-02$ & $1e-02$ & $0e-00$ & $1e-02$ & $1e-02$ & $6e-02$ \\ 
&  Maximum Sequence Lengt & \multicolumn{8}{c}{$256$} \\
\Xhline{1pt}
\end{tabular}
\caption{Hyperparameter values for training T5 base on the GLUE benchmark.}
\label{tab:hyper_appendix_t5_base}
\end{table*}

\begin{table*}[ht] 
\small
\centering
\setlength{\tabcolsep}{3pt}
\begin{tabular}{l|l|cccccccc}
\Xhline{1pt}
\textbf{Method} & \textbf{Dataset} & \textbf{MNLI} & \textbf{SST-2} & \textbf{MRPC} & \textbf{CoLA} & \textbf{QNLI} & \textbf{QQP} & \textbf{RTE} & \textbf{STS-B} \\
\Xhline{1pt}
\multirow{2}{*}{\textbf{ALL}} & {Optimizer} &  \multicolumn{8}{c}{AdamW}\\
& LR Schedule &  \multicolumn{8}{c}{Linear}\\
\hline
\multirow{5}{*}{\textbf{FT}} & Batch Size & $16$ & $32$ & $32$ & $32$ & $16$ & $16$ & $16$ & $32$ \\
& Number of Epochs & $10$ & $10$ & $20$ & $20$ & $10$ & $10$ & $20$ & $10$ \\
&  Learning rate & $2e-05$ & $2e-05$ & $2e-05$ & $2e-05$ & $2e-05$ & $1e-05$ & $1e-05$ & $2e-05$ \\ 
& Warmup Ratio & $6e-02$ & $6e-02$ & $6e-02$ & $6e-02$ & $6e-02$ & $6e-02$ & $6e-02$ & $6e-02$ \\ 
&  Weight Decay & \multicolumn{8}{c}{$1e-01$} \\
&  Maximum Sequence Length & \multicolumn{8}{c}{$256$} \\
\hline
\multirow{5}{*}{\textbf{LoRA}} & Batch Size & $32$ & $32$ & $32$ & $32$ & $32$ & $32$ & $32$ & $32$ \\
& Number of Epochs & $10$ & $10$ & $20$ & $20$ & $10$ & $10$ & $20$ & $10$ \\
&  Learning rate & $3e-04$ & $4e-04$ & $3e-04$ & $2e-04$ & $2e-04$ & $3e-04$ & $4e-04$ & $2e-04$ \\
 & Warmup Ratio & $6e-02$ & $6e-02$ & $6e-02$ & $6e-02$ & $6e-02$ & $6e-02$ & $6e-02$ & $6e-02$ \\ 
& Rank & \multicolumn{8}{c}{$8$} \\
& Coefficient $\alpha$ & \multicolumn{8}{c}{$16$} \\
& Maximum Sequence Length & \multicolumn{8}{c}{$256$} \\
\hline
\multirow{5}{*}{\textbf{Adapter}} & Batch Size & $32$ & $32$ & $32$ & $32$ & $32$ & $32$ & $32$ & $32$ \\
& Number of Epochs & $10$ & $10$ & $20$ & $20$ & $10$ & $10$ & $20$ & $10$ \\
&  Learning rate & $3e-04$ & $3e-04$ & $3e-04$ & $3e-04$ & $3e-04$ & $3e-04$ & $3e-04$ & $3e-04$ \\
& Warmup Ratio & $6e-02$ & $6e-02$ & $6e-02$ & $6e-02$ & $6e-02$ & $6e-02$ & $6e-02$ & $6e-02$ \\
& Rank & \multicolumn{8}{c}{$8$} \\
&  Maximum Sequence Length & \multicolumn{8}{c}{$256$} \\
\hline
\multirow{5}{*}{\textbf{Adapter-FFN}} & Batch Size & $32$ & $32$ & $32$ & $32$ & $32$ & $32$ & $32$ & $32$ \\
& Number of Epochs & $10$ & $10$ & $20$ & $20$ & $10$ & $10$ & $20$ & $10$ \\
&  Learning rate & $3e-04$ & $3e-04$ & $3e-04$ & $3e-04$ & $3e-04$ & $3e-04$ & $3e-04$ & $3e-04$ \\
& Warmup Ratio & $6e-02$ & $6e-02$ & $6e-02$ & $6e-02$ & $6e-02$ & $6e-02$ & $6e-02$ & $6e-02$ \\
& Rank & \multicolumn{8}{c}{$16$} \\
&  Maximum Sequence Length & \multicolumn{8}{c}{$256$} \\
\hline
\multirow{5}{*}{\textbf{RED}} & Batch Size & $32$ & $32$ & $32$ & $32$ & $32$ & $32$ & $32$ & $32$ \\
& Number of Epochs & $10$ & $10$ & $20$ & $20$ & $10$ & $10$ & $20$ & $10$ \\
 &  Learning rate & $1e-03$ & $1e-03$ & $2e-03$ & $1e-03$ & $1e-03$ & $1e-03$ & $5e-03$ & $5e-03$ \\
& Weight Decay & $0.0$ & $0.0$ & $0.0$ & $0.0$ & $0.0$ & $0.0$ & $1e-04$ & $0.0$ \\
& Warmup Ratio & $6e-02$ & $6e-02$ & $0$ & $6e-02$ & $6e-02$ & $6e-02$ & $1e-02$ & $6e-02$ \\
&  Maximum Sequence Length & \multicolumn{8}{c}{$256$} \\
\Xhline{1pt}
\end{tabular}
\caption{Hyperparameter values for training RoBERTa large on the GLUE benchmark.}
\label{tab:hyper_appendix_roberta_large}
\end{table*}

\begin{table*}[ht] 
\small
\centering
\setlength{\tabcolsep}{2pt}
\begin{tabular}{l|cccccccc}
\Xhline{1pt}
\textbf{Dataset} & \textbf{FT} & \textbf{$\mathbf{FT^{top2}}$} & \textbf{Adapter} & \textbf{Apapter-FFN} & \textbf{LoRA} & \textbf{Prefix Tuning} & \textbf{RED (medium)} & \textbf{RED (large)} \\
\Xhline{1pt} 
\multicolumn{9}{c}{\textbf{Training}} \\
\Xhline{1pt}
Optimizer & \multicolumn{8}{c}{AdamW} \\
Weight Decay & $0.0$ & $0.0$ & $0.0$ & $0.0$ & $1e-02$ & $0.0$ & $1e-04$ & $0.0$\\
Number of Epochs & $5$ & $5$ & $5$ & $5$ & $5$ & $5$ & $5$ & $10$\\
Learning Rate Schedule & \multicolumn{8}{c}{Linear} \\
Label Smooth & $0.0$ & $0.0$ & $0.0$ & $0.0$ & $0.1$ & $0.0$ & $0.0$ & $0.0$\\
Learning Rate & $5e-05$ & $5e-05$ & $8e-05$ & $8e-05$ & $2e-04$ & $8e-05$ & $6e-02$ & $6e-03$\\
Rank or Prefix Length & $--$ & $--$ & $8$ & $16$ & $8$ & $16$ & $--$ & $--$ \\
Coefficient $\alpha$ & $--$ & $--$ & $--$ & $--$ & $32$ & $--$ & $--$ & $--$ \\
Adaption & $--$ & $--$ & $--$ & $--$ & $8$ & $--$ & $--$ & $--$ \\
Warmup Steps & \multicolumn{8}{c}{$500$} \\
Batch Size & \multicolumn{8}{c}{$10$} \\
\Xhline{1pt}
\multicolumn{9}{c}{\textbf{Inference}} \\
\Xhline{1pt}
Beam Size & \multicolumn{8}{c}{$10$} \\
Length Penalty & \multicolumn{8}{c}{$0.9$}\\
No-repeat $n$-gram Size & \multicolumn{8}{c}{$4$}\\
\Xhline{1pt}
\end{tabular}
\caption{Hyperparameter values for training GPT-2 on the E2E NLG Challenge. ``RED (medium)'' denotes the values of hyperparameters used by RED to fine-tune GPT-2 medium and ``RED (large)'' the values to GPT-2 large.}
\label{tab:hyper_appendix_gpt2}
\end{table*}

\begin{table*}[ht] 
\small
\centering
\setlength{\tabcolsep}{1.2pt}
\begin{tabular}{l|r|cccccccc|c}
\Xhline{1pt}
\textbf{Methods} & \textbf{\# Param} & \textbf{MNLI} & \textbf{SST-2} & \textbf{MRPC} & \textbf{CoLA} & \textbf{QNLI} & \textbf{QQP} & \textbf{RTE} & \textbf{STS-B} & \textbf{Avg.} \\
\Xhline{1pt}
\textbf{FT (base)} & $125\mathbf{M}$ & $87.3_{\small \pm 0.34}$ & $94.4_{\small \pm 0.96}$ & $87.9_{\small \pm 0.91}$ & $62.4_{\small \pm 3.29}$ & $92.5_{\small \pm 0.22}$ & $91.7_{\small \pm 0.19}$ & $78.3_{\small \pm 3.20}$ & $90.6_{\small \pm 0.59}$ & $85.6$\\

\textbf{Adapter (base)} & $0.4\mathbf{M}$ & $87.0_{\small \pm 0.28}$ & $93.3_{\small \pm 0.40}$ & $88.4_{\small \pm 1.54}$ & $60.9_{\small \pm 3.09}$ & $92.5_{\small \pm 0.02}$ & $90.5_{\small \pm 0.08}$ & $76.6_{\small \pm 2.26}$ & $90.5_{\small \pm 0.35}$ & $85.0$ \\

\textbf{Adapter-FFN (base)} & $0.3\mathbf{M}$ & $87.1_{\small \pm 0.10}$ & $93.0_{\small \pm 0.50}$ & $88.8_{\small \pm 1.38}$ & $58.5_{\small \pm 1.69}$ & $92.1_{\small \pm 0.28}$ & $90.2_{\small \pm 0.07}$ & $77.7_{\small \pm 1.93}$ & $90.4_{\small \pm 0.31}$ & $84.7$ \\

\textbf{LoRA (base)} & $0.3\mathbf{M}$ & $86.6_{\small \pm 0.26}$ & $93.9_{\small \pm 0.49}$ & $88.7_{\small \pm 0.76}$ & $59.7_{\small \pm 4.36}$ & $92.6_{\small \pm 0.10}$ & $90.4_{\small \pm 0.08}$ & $75.3_{\small \pm 2.79}$ & $90.3_{\small \pm 0.54}$ & $84.7$ \\

\textbf{BitFit (base)} & $0.1\mathbf{M}$ & $84.7_{\small \pm 0.08}$ & $94.0_{\small \pm 0.87}$ & $88.1_{\small \pm 1.57}$ & $54.0_{\small \pm 3.07}$ & $91.0_{\small \pm 0.05}$ & $87.3_{\small \pm 0.02}$ & $69.8_{\small \pm 1.51}$ & $89.5_{\small \pm 0.35}$ & $82.3$ \\

\rowcolor{gray!30}
\textbf{RED (base)} & $0.02\mathbf{M}$ & $83.9_{\small \pm 0.14}$ & $93.9_{\small \pm 0.31}$ & $89.2_{\small \pm 0.98}$ & $61.0_{\small \pm 2.96}$ & $90.7_{\small \pm 0.35}$ & $87.2_{\small \pm 0.17}$ & $78.0_{\small \pm 2.06}$ & $90.4_{\small \pm 0.32}$ & $84.7$ \\

\hline
\textbf{FT (large)} & $355\mathbf{M}$ & $88.8_{\small \pm 0.45}$ & $96.0_{\small \pm 0.66}$ & $91.7_{\small \pm 1.73}$ & $68.2_{\small \pm 2.62}$ & $93.8_{\small \pm 0.33}$ & $91.5_{\small \pm 1.28}$ & $85.8_{\small \pm 1.40}$ & $92.6_{\small \pm 0.16}$ & $88.5$ \\

\textbf{LoRA (large)} & $0.8\mathbf{M}$ & $90.2_{\small \pm 0.25}$ & $96.0_{\small \pm 0.85}$ & $89.8_{\small \pm 2.09}$ & $65.5_{\small \pm 2.02}$ & $94.7_{\small \pm 0.21}$ & $90.7_{\small \pm 0.91}$ & $86.3_{\small \pm 2.41}$ & $91.7_{\small \pm 0.44}$ & $88.1$ \\

\textbf{Adapter (large)} & $0.9\mathbf{M}$ & $90.1_{\small \pm 0.12}$ & $95.2_{\small \pm 0.48}$ & $90.5_{\small \pm 0.59}$ & $65.4_{\small \pm 2.24}$ & $94.6_{\small \pm 0.17}$ & $91.4_{\small \pm 0.13}$ & $85.3_{\small \pm 1.34}$ & $91.5_{\small \pm 0.33}$ & $88.0$ \\

\textbf{Adapter-FFN (large)} & $0.8\mathbf{M}$ & $90.3_{\small \pm 0.15}$ & $96.1_{\small \pm 0.75}$ & $90.5_{\small \pm 1.26}$ & $64.4_{\small \pm 1.56}$ & $94.3_{\small \pm 0.39}$ & $91.3_{\small \pm 0.24}$ & $84.8_{\small \pm 2.01}$ & $90.2_{\small \pm 0.24}$ & $87.7$ \\

\rowcolor{gray!30}
\textbf{RED (large)} & $0.05\mathbf{M}$ & $89.5_{\small \pm 0.38}$ & $96.0_{\small \pm 0.48}$ & $90.3_{\small \pm 1.40}$ & $68.1_{\small \pm 1.69}$ & $93.5_{\small \pm 0.33}$ & $88.8_{\small \pm 0.11}$ & $86.2_{\small \pm 1.40}$ & $91.3_{\small \pm 0.21}$ & $87.9$ \\
\Xhline{1pt}
\end{tabular}
\caption{Performance comparison of RoBERTa base and large models fine-tuned by RED and other PEFT baselines on the GLUE benchmark.}
\label{tab:standaerd_appendix_roberta}
\end{table*}

\begin{table*}[ht] 
\small
\centering
\setlength{\tabcolsep}{8.5pt}
\begin{tabular}{l|r|ccccc}
\Xhline{1pt}
\textbf{Method} & \textbf{\# Param} & \textbf{BLEU} & \textbf{NIST} & \textbf{MET} & \textbf{ROUGE-L} & \textbf{CIDEr} \\
\Xhline{1pt}
\textbf{FT (medium)} & $355\mathbf{M}$ & $65.95_{\small \pm 0.26}$ & $8.52_{\small \pm 0.03}$ & $45.95_{\small \pm 0.07}$ & $69.13_{\small \pm 0.30}$ & $2.35_{\small \pm 0.01}$  \\
\textbf{$\mathbf{FT^{top2}}$ (medium)} & $25.2\mathbf{M}$ & $65.94_{\small \pm 0.33}$ & $8.53_{\small \pm 0.03}$ & $44.28_{\small \pm 0.09}$ & $68.83_{\small \pm 0.17}$ & $2.23_{\small \pm 0.02}$  \\

\textbf{Adapter (medium)} & $0.9\mathbf{M}$ & $64.31_{\small \pm 0.17}$ & $8.29_{\small \pm 0.01}$ & $44.91_{\small \pm 0.29}$ & $67.72_{\small \pm 0.26}$ & $2.28_{\small \pm 0.01}$  \\

\textbf{Adapter-FFN (medium)} & $0.8\mathbf{M}$ & $64.41_{\small \pm 0.17}$ & $8.30_{\small \pm 0.02}$ & $44.74_{\small \pm 0.11}$ & $67.53_{\small \pm 0.02}$ & $2.29_{\small \pm 0.01}$ \\

\textbf{LoRA (medium)} & $0.8\mathbf{M}$ & $67.43_{\small \pm 0.39}$ & $8.65_{\small \pm 0.05}$ & $46.01_{\small \pm 0.07}$ & $69.64_{\small \pm 0.14}$ & $2.42_{\small \pm 0.01}$  \\

\textbf{Prefix Tuning (medium)} & $0.8\mathbf{M}$ & $63.92_{\small \pm 0.27}$ & $8.26_{\small \pm 0.11}$ & $41.81_{\small \pm 0.62}$ & $66.86_{\small \pm 0.22}$ & $2.03_{\small \pm 0.05}$  \\

\rowcolor{gray!30}
\textbf{RED (medium)} & $0.05\mathbf{M}$ & $64.86_{\small \pm 0.40}$ & $8.36_{\small \pm 0.03}$ & $44.99_{\small \pm 0.02}$ & $67.62_{\small \pm 0.22}$ & $2.28_{\small \pm 0.01}$  \\

\hline 
\textbf{FT (large)} & $774\mathbf{M}$ & $65.56_{\small \pm 0.47}$ & $8.50_{\small \pm 0.05}$ & $45.40_{\small \pm 0.29}$ & $68.38_{\small \pm 0.23}$ & $2.27_{\small \pm 0.02}$  \\

\textbf{Adapter (large)} & $1.8\mathbf{M}$ & $65.94_{\small \pm 0.35}$ & $8.46_{\small \pm 0.05}$ & $45.78_{\small \pm 0.11}$ & $68.65_{\small \pm 0.35}$ & $2.34_{\small \pm 0.01}$  \\

\textbf{Adapter-FFN (large)} &$1.5\mathbf{M}$ & $65.53_{\small \pm 0.61}$ & $8.41_{\small \pm 0.07}$ & $45.65_{\small \pm 0.12}$ & $68.46_{\small \pm 0.16}$ & $2.33_{\small \pm 0.01}$  \\

\textbf{LoRA (large)} & $1.5\mathbf{M}$ & $68.24_{\small \pm 0.28}$ & $8.76_{\small \pm 0.04}$ & $46.23_{\small \pm 0.04}$ & $69.92_{\small \pm 0.16}$ & $2.42_{\small \pm 0.01}$  \\

\textbf{Prefix Tuning (large)} & $1.5\mathbf{M}$ & $65.50_{\small \pm 0.63}$ & $8.45_{\small \pm 0.05}$ & $43.97_{\small \pm 0.21}$ & $67.32_{\small \pm 0.38}$ & $2.23_{\small \pm 0.02}$  \\

\rowcolor{gray!30}
\textbf{RED (large)} & $0.09\mathbf{M}$ & $65.77_{\small \pm 0.48}$ & $8.42_{\small \pm 0.06}$ & $46.12_{\small \pm 0.10}$ & $69.03_{\small \pm 0.09}$ & $2.36_{\small \pm 0.02}$  \\

\Xhline{1pt}
\end{tabular}
\caption{Performance comparison of GPT-2 medium and large models fine-tuned by RED and other PEFT baselines on the E2E NLG Challenge.}
\label{tab:standaerd_appendix_gpt2}
\end{table*}

\begin{table*}[ht]
\small
\centering
\setlength{\tabcolsep}{4pt}
\begin{tabular}{l|c|cccccccc|c}
\Xhline{1pt} 
\textbf{Method} &      \textbf{\# Param}   & \textbf{Writing}     & \textbf{Roleplay} & \textbf{Reasoning}    & \textbf{Math} & \textbf{Coding} & \textbf{Extraction} & \textbf{Stem}  & \textbf{Humanities} & \textbf{Average}\\
\hline
\multicolumn{1}{l}{\textbf{Turn-1}} \\
\hline
\textbf{FT}  &$6739\mathbf{M}$& $9.111$ & $6.500$   & $4.778$ & $1.444$  &  $2.700$  & $4.800$   & $7.944$ & $8.833$ & $5.688$  \\
\textbf{LoRA} &$8.39\mathbf{M}$ & $8.600$ & $7.900$   & $4.000$ & $1.600$  &  $2.350$  & $4.750$  & $8.700$ & $8.950$ & $5.856$ \\
\rowcolor{gray!30}
\textbf{RED}  &$0.26\mathbf{M}$ &     $8.778$    & $7.278$   &    $4.875$     &      $1.300$  &  $2.625$  & $4.125$    &   $7.400$    & $9.167$ & $5.732$   \\
\hline
\multicolumn{1}{l}{\textbf{Turn-2}} \\
\hline
\textbf{FT}  & $6739\mathbf{M}$ & $1.667$ & $5.938$   & $2.222$ & $1.700$  &  $2.000$  & $2.111$   & $3.200$ & $5.300$ & $3.021$  \\
\textbf{LoRA}  &$8.39\mathbf{M}$ & $1.900$ & $5.800$   & $2.100$ & $1.600$  &  $2.550$  & $1.222$  & $3.100$ & $5.500$ & $2.994$ \\
\rowcolor{gray!30}
\textbf{RED}  &$0.26\mathbf{M}$ &     $1.375$    & $5.500$   &    $2.444$     &      $1.444$  &  $2.125$  & $1.750$    &   $3.000$    & $5.875$ & $2.946$   \\
\hline
\multicolumn{1}{l}{\textbf{Average}} \\
\hline
\textbf{FT} &$6739\mathbf{M}$ & $5.389$ & $6.200$   & $3.500$ & $1.579$  &  $2.412$  & $3.526$   & $5.447$ & $6.974$ & $4.354$  \\
\textbf{LoRA} &$8.39\mathbf{M}$ & $5.250$ & $6.850$   & $3.050$ & $1.600$  &  $2.450$  & $3.079$  & $5.900$ & $7.225$ & $4.434$ \\
\rowcolor{gray!30}
\textbf{RED} &$0.26\mathbf{M}$ &     $5.294$    & $6.389$   &    $3.588$     &      $1.368$  &  $2.375$  & $2.938$    &   $5.750$    & $7.618$ & $4.401$   \\
\Xhline{1pt} 
\end{tabular}
\caption{Performance comparison of LLaMA-2 fine-tuned by RED and other PEFT baselines on the MT-Bench.
}
\label{tab:mt-details}   
\end{table*}


\end{document}